\documentclass[12pt,letterpaper]{article}
\usepackage[a4paper, total={7in, 10in}]{geometry}

\usepackage{graphicx}
\usepackage{helvet}
\usepackage{authblk}
\usepackage{hyperref}
\usepackage{amsmath} 
\usepackage{amssymb} 
\usepackage{orcidlink} 
\usepackage[super,comma,sort&compress]  
   {natbib}

\usepackage{caption}
\usepackage{subcaption}

\makeatletter
\renewcommand{\maketitle}{\bgroup\setlength{\parindent}{0pt}
\begin{flushleft}
  \textbf{\@title}
  
  \@author
\end{flushleft}\egroup}
\makeatother

\title{A Physics-Flavored Transformer Network for Parametrizing Contraction Dynamics of Engineered Skeletal Muscle Tissues}
\date{}

\author[1,2,\orcidlink{0000-0002-3775-0804}]{Mattias Luber}
\author[2,3,*]{Timo Betz}

\affil[1]{CIDAS: Campus-Institute Data Science, Goldschmidtstraße 1,
Göttingen, 37077, Lower Saxony, Germany}
\affil[2]{Third Institute of Physics, University of Göttingen,
Friedrich-Hund-Platz 1, Göttingen, 37077, Lower Saxony,
Germany}

\affil[3]{Lead contact}

\affil[*]{Correspondence: timo.betz@phys.uni-goettingen.de}

\begin{document}

\maketitle

\section*{SUMMARY}

Engineered Skeletal Muscle Tissues (ESMs) have become a key structure for biomedical disease modeling and pharmacological screening, yet their functional characterization often relies on simplistic metrics like peak force, discarding critical kinetic information. This is partially due to the high level of mathematical complexity which mechanistic models introduce to capture these dynamics. Hence, exactly the complexity prevents scalable application and widespread adaptation in the field. Here we present a Physics-Flavored Neural Network (PFNN) that automates the kinetic phenotyping of ESMs. Our architecture integrates a stretched-exponential physical model into a CNN-Transformer, enabling the extraction of physically meaningful parameters directly from force-time profiles. To address the scarcity of labeled biological data, we employ a hybrid training paradigm: the model develops a "physical intuition" on synthetic data before undergoing unsupervised self-alignment on unlabeled real-world measurements. Our results demonstrate that this physics-flavored approach achieves high-fidelity parameterization across diverse contractile phenotypes and cell lines, including Duchenne Muscular Dystrophy models. Our scalable, self-improving pipeline bridges the gap between idealized biophysics and noisy \emph{in vitro} data, providing a robust tool for high-throughput biophysical research.

\section*{KEYWORDS}

Deep Learning, Physics-Informed Neural Network, Tissue Engineering, Skeletal Muscle 

\section{Introduction}\label{sec1}
Engineered Skeletal Muscle Tissues (ESM) have emerged as a crucial translational tool for modern biomedical research, providing a high-fidelity \emph{in vitro} platform to investigate muscle development\cite{AfsharBakooshli.2019} or to develop gene therapies for neuromuscular diseases such as Duchenne Muscular Dystrophy (DMD) \cite{Shahriyari.2022}. By mimicking the structural and functional complexity of native tissue within a highly controlled environment, ESMs serve as a sophisticated biological testbed for evaluating pharmacological treatments \cite{Yoshida.2025} or investigating muscle aging and regeneration \cite{Rajabian.2021}. Central to these studies is the quantification of contractile force, which is the main functional readout for many translational experiments \cite{VesgaCastro.2022,Moyle.2020,vanderWal.2023}.

Intracellularly, these contractions are the product of a complex, time-dependent cascade of coupled actuation and transmission processes. An electrical stimulus triggers a rapid release of calcium ions, activating an ensemble of molecular motor proteins that collectively drive the shortening of individual muscle fibers. This active force is coupled via transmembrane proteins to the surrounding extracellular matrix (ECM), where it is further modulated by the matrix's passive viscoelastic properties.\cite{Rohrle.2019} The final force-time profile is thus an emergent signature of both active molecular kinetics and passive structural resistance. In many pathological states, specific components of this machinery, such as motor protein recruitment or ECM stiffness, become altered or misregulated \cite{Lieber.2013,Hill.2021,Wang.2025b}. These changes manifest as characteristic shifts in the temporal evolution of the contraction curve, impacting its shape and kinetic constants. 

The quantification of muscle kinetics has traditionally followed two divergent paths: mechanistic and phenomenological \cite{Wakeling.2023}. Mechanistic models, most notably the Huxley cross-bridge model\cite{HUXLEY.1954}, were originally designed to explain force generation through the interplay of molecular filaments at the sarcomere scale. While efforts have been made to scale these frameworks up to muscle fibers or entire muscles e.g. via multiscale or continuum muscle models \cite{Oomens.2003,Milicevic.2022,VillotaNarvaez.2022,Zeng.2023}, those approaches are rarely used in physiological experiments. This is partially due to their (partial) differential equation structure \cite{Zahalak.1981,vanSoest.2019}, high dimensionality, and the inherent difficulty of experimentally accessing their numerous parameters like e.g. molecular binding rates \cite{Zeng.2023,Wang.2025,Heidlauf.2016,HernandezGascon.2013,Karami.2023,Lemaire.2016}. Consequently, researchers often default to simple heuristic observables, such as peak force, time-to-peak or half-relaxation time, to quantify contractile phenotypes \cite{Dennis.2001,Shahriyari.2022,VesgaCastro.2022}. While accessible, these discrete metrics fail to capture the full continuous temporal evolution of the contraction, discarding valuable information that could be used for assessing the efficiency of therapies.

To address these limitations, we previously proposed a phenomenological framework based on stretched exponential functions and demonstrated its capability to accurately capture contraction dynamics across a wide range of pharmacological conditions with a minimal set of intuitive parameters \cite{Luber.2026}. However, the fitting of these parameters relied on classical trust-region optimization, which crucially requires manual derivation of suitable initial guesses for the fit parameters to avoid that the fitting routine remains stuck in local minima. This manual interaction prevents the method from scaling to large-scale screening.

In contrast deep learning models are generally considered to be scaleable and furthermore allow the intragration of prior physical knowledge. The general concept of the integration of physical laws into neural networks has been formalized by \citeauthor{Raissi.2019} as Physics inspired neural networks to model (partial) differential equations, solve inverse problems or improve the performance of neural networks on small empirical datasets. Over the years PINNs have been shown to be successful for a multitude of tasks and in a broad set of research fields like in fluid mechanics \cite{Cai.2021}, astrophysics \cite{Janssen.2025} or in biomedical engineering\cite{Ahmadi.2026}.

In the present work, we employ a CNN-Transformer architecture to estimate physically meaningful parameters from \emph{in vitro} skeletal muscle tissue contractions, with a focus on the tetanus contractions. We utilize a relatively slim, physics-grounded model equation in a dual-stage process. First, it is used to generate synthetic data that mimics the appearance of our real measurement data, thereby overcoming the scarcity of labeled biological samples. This is conceptually similar to other PINN-approaches that generate artificial data and integrate it into the training pipeline to improve model performance \cite{Jeong.2025,GuerreroViu.2021,Wang.2025c}. Second, the model equation serves as a physics-based loss function to assess the reconstruction error for both synthetic and unlabeled real-world data, based on the parameters predicted by the network. Due to this dual integration of data synthesis and loss regularization, we term our approach Physics-Flavored Neural Networks (PFNN).

Our approach adopts a design principle similar to the vision transformer \cite{Dosovitskiy.2020}. Specifically, we utilize a series of shallow CNN layers for learned preprocessing and initial feature extraction. These features are then projected into an embedding space, augmented with positional encodings, and processed by Transformer Encoder layers to aggregate global information into final parameter predictions for the tetanus contractions via specialized tokens in the latent space.  \cite{Devlin.2018,Dosovitskiy.2020}. By mapping these parameters into an unconstrained optimization space, we integrate our physical model directly into the network as a physics-flavored loss function.

The model is trained primarily on an extensive library of synthetic data generated from the physical model with known ground-truth parameters to develop an initial "physical intuition". Additionally, we inject unlabeled real-world measurements during the training using an unsupervised reconstruction loss. We demonstrate that this physics-flavored approach yields excellent fit quality and provides highly accurate parameter predictions. Crucially, the injection of real measurement data without the necessity of human-provided labels during training, establishing a self-improving system, that scales efficiently with new incoming data. By utilizing these neural predictions as high-quality initial values for a final trust-region optimization, new labels can be generated, creating a self-reinforcing, automated analysis pipeline. This framework offers a scalable, objective, and robust solution for the deep phenotyping of engineered muscle tissues.

\section{Methods}
\subsection{Data description}

\begin{figure}
    \centering
    \includegraphics[width=0.99\linewidth]{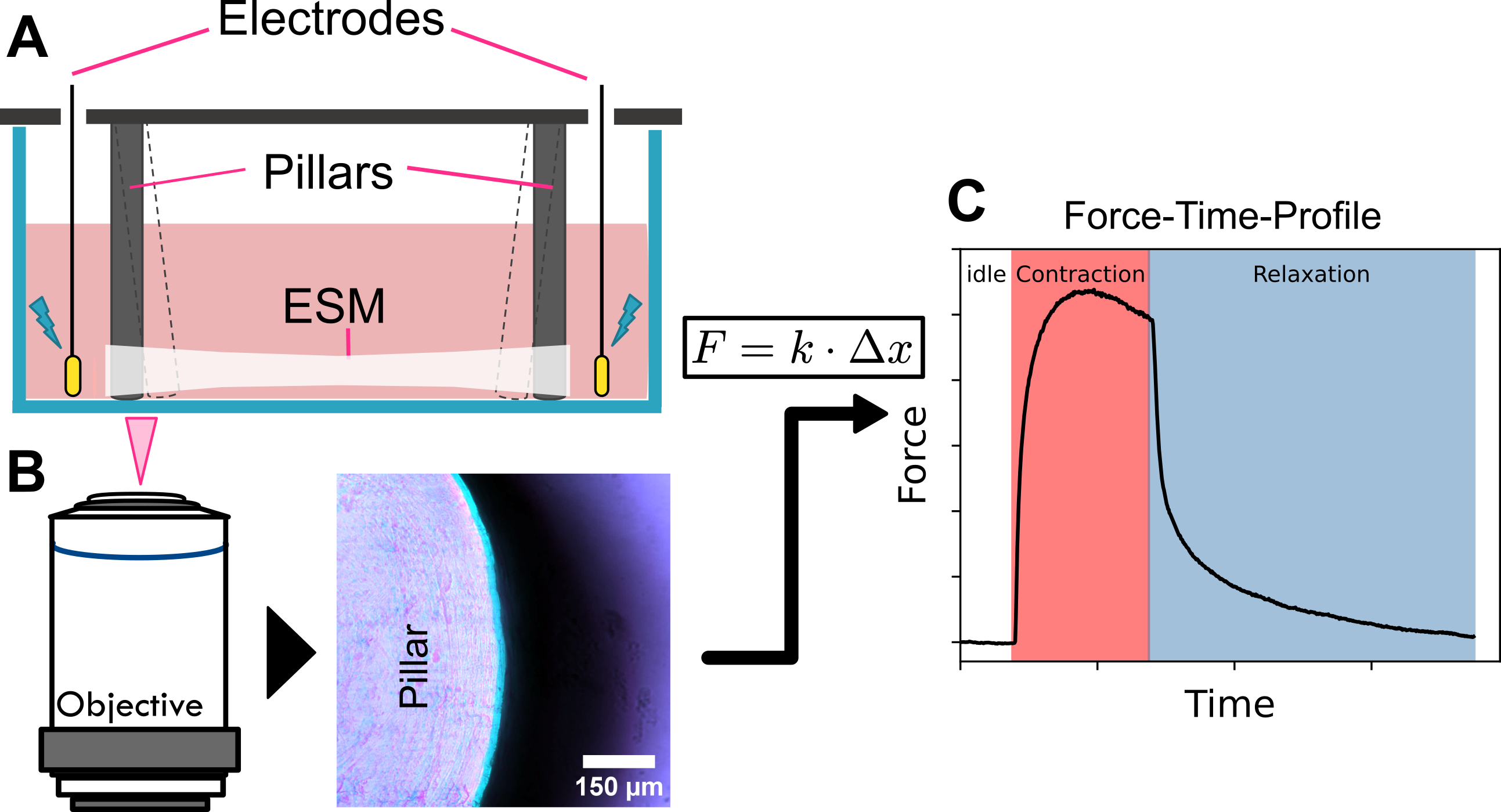}
    \caption{Sketch of the experimental data derivation for the force-time profiles of engineered skeletal muscle tissues: A) Tissues are cultured in specialized dishes between two flexible, elastic pillars. Electrical stimulation via steel electrodes over four several seconds induces sustained contractions (tetani). B) As the muscle tissue contracts, it exerts a force on the pillars, leading to a micrometer-scale deflection (highlighted in cyan). This deflection is captured via microscopy video recordings. Utilizing Hooke’s Law and the pre-calibrated spring constant of the pillars, the displacement is converted into force-time profiles. C) These tetanus contractions exhibit a Force-Time profile with a characteristic unimodal shape, divisible into an initial idle phase (pre-stimulation), a contraction phase reaching peak force and a subsequent relaxation phase upon cessation of the electrical stimulus.}
    \label{fig:OverviewDataGeneration}
\end{figure}
Different approaches exist to quantify contractile forces of ESMs e.g. directly by employing force transducers or indirectly by using elastic pillars as force sensors \cite{VesgaCastro.2022}. The data in this work is obtained by a previously published pillar-based system \cite{Hofemeier.2021}. Tissues are raised in attachment to two elastic pillars and contractions are triggered via electrical stimulation over a duration of approximately 4 seconds (\ref{fig:OverviewDataGeneration}A). Upon contraction the muscle tissues exerts tension on the flexible pillars, resulting in a micrometer-scale displacement. By analyzing microscopy video recordings, these contractions can be quantified (\ref{fig:OverviewDataGeneration}B). The mechanical force is derived from the measured pillar deflection $\Delta x$ using Hooke’s Law ($F=k\cdot \Delta x$) where $k$ represents the known spring constant of the pillar. The experimental procedure and the resulting characteristic force-time profiles are illustrated in Figure \ref{fig:OverviewDataGeneration}C.

The resulting force-time profiles of a tetanus contraction follow a characteristic triphasic temporal evolution: an initial rise phase towards the peak force, a plateau phase representing sustained contraction, and a subsequent decay upon the cessation of the stimulus. In pathological or immature tissues, this stereotypical progression is frequently altered. Deviations in specific kinetic features such as diminished peak force, prolonged relaxation times, or plateau instability due to fatigue, serve as critical quantitative indicators for characterizing disease models or evaluating the efficacy of pharmacological interventions \cite{Madden.2015, Smith.2022, Shahriyari.2022,Luber.2026}.

Our experimental foundation consists of two distinct datasets. The primary dataset, derived from our previous work \cite{Luber.2026}, comprises 370 tetanus contractions of ESMs derived from the LHCN-M2 cell line. These contractions were recorded at a frame rate of 50 fps over a duration of 15 seconds, yielding 750 frames per contraction curve. The ESMs in this dataset were subjected to a library of pharmacological compounds that affected different components of the contractile machinery, including calcium handling, motor protein recruitment, and cross-bridge cycling dynamics. Thus, these compounds induce a broad spectrum of contractile phenotypes representative under various biomedical conditions. This dataset is used as a primary training and validation set for this study.

To evaluate the model's generalization across a broader biological context, we incorporated a secondary test dataset into this study. While obtained under similar experimental conditions, this cohort originates from three biologically distinct human donor cell lines (AB1167, KM571, KM670 - Institute of Myology at the Pitié-Salpêtrière hospital in Paris), introducing profound physiological variance across age, anatomical origin, and disease state. Specifically, the healthy control line AB1167 originates from the \emph{fascia lata} of a 20-year-old male and has been well-characterized for its robust functional contractility in 3D tissue-engineered platforms \cite{Tollitt.2025,Tiper.2025,Hofemeier.2021,Hofemeier.2022}. To demonstrate the platform's utility in disease modeling, the KM571 cell line was included. To model pathological states where the contractile machinery is functionally altered, the KM571 line derived from the same anatomical origin. The cell line originates from a 13-year-old male Duchenne Muscular Dystrophy (DMD) patient harboring exon 45–52 and exon 52 deletions and thus represents the physiological hallmarks of dystrophin deficiency. Finally, the KM670 cell line, isolated from the \emph{quadriceps} of an 83-year-old male donor, introduces age-associated myogenic variance. This dataset represents an independent hold-out test set.

Both datasets are unlabeled in the context of supervised deep learning. Since the true underlying kinetic parameters governing these biological curves are not directly observable, no experimental ground truth exists for direct model supervision.

\subsection{Theory and background}
To characterize these complex kinetics of tetanic contraction and subsequent fatigue in \emph{in-vitro} skeletal muscle, we employ a phenomenological model based on the Kohlrausch-Williams-Watts \cite{Kohlrausch.1854}(KWW) stretched exponential function. While we have previously demonstrated the utility of this model in pharmacological drug testing \cite{Luber.2026}, here we motivate its mathematical necessity through the biophysical reality of tissue heterogeneity and the data-centric requirement for parameter identification.

From a biophysical standpoint, on a single fiber level a muscle contraction is an emergent result of coupled, rate-dependent processes interacting across disparate timescales from millisecond-scale $Ca^{2+}$ handling to delayed viscoelastic response of the extra-cellular-matrix\cite{Rausch.2020}. Furthermore, an engineered tissue construct consists of thousands of individual fibers, each in a potentially different state of differentiation, structural alignment, or motor protein distribution.\cite{MorenoJusticia.2025,Juhas.2015,Cheng.2014,Luber.2026}

Consequently, the collective contractile response does not operate on a single, well-defined time scale. While a naive approach might attempt to model this via a discrete sum of exponentials, $F(t)= \sum A_i e^{\frac{-t}{\tau_i}}$, this inverse problem is notoriously ill-posed\cite{Istratov.1999}. Resolving the specific number of components or even the full spectrum is non-obvious, and the solution is highly sensitive to experimental noise, often requiring regularization that add additional layers of complexity and hence bias the biological interpretation. \cite{Istratov.1999,Spencer.2020,Sabett.2017}

To circumvent these limitations, we treat the muscle construct as a complex system defined by a continuous distribution of relaxation rates. The mathematical beauty of the stretched exponential lies in its representation as a weighted integral of a density function $\rho(u)$\cite{Elton.2018} for some non-trivial $\rho(u)$\cite{BerberanSantos.2005,Lindsey.1980,Johnston.2006}:

\begin{equation}
    F(t) = A_0 \exp\left( -\left( \frac{t}{\tau} \right)^\beta \right) = \int_0^\infty \rho(u) e^{-t/u} du
\end{equation}

By adopting this form, we acknowledge the intrinsic heterogeneity in timescales of the underlying biological assembly without attempting to isolate individual molecular timescales that are practically unidentifiable. In this framework, the parameters carry deep physical meaning:
\begin{itemize}
    \item $\tau$ Represents the effective, aggregate timescale of the process. 
    \item The stretching exponent $\beta$: Where $0<\beta\leq1$, serves as a direct measure of the system’s heterogeneity. A $\beta=1$ indicates a perfectly synchronized system (a single exponential), while a decreasing $\beta$ quantifies the broadening distribution of underlying rates.
\end{itemize}

By separating the kinetics into these well-defined and interpretable parameters, the model becomes especially appealing for biomedical applications. It allows one to quantify the full contraction curve jointly into a small set of intuitive parameters that directly allow to e.g. reason about whether a drug treatment improves the overall speed of contraction (shifting $\tau$) or to quantify the synchronization of fiber recruitment (shifting $\beta$). 
From a physics point of view the stretched exponential compensates for deviation from idealized purely exponential decay functions as they are common in real-world application and effectively models a system where energy dissipation changes as the system relaxes \cite{Lukichev.2019}. In practice, stretched exponentials have a broad range of applications and were previously used to describe e.g. fluorescent lifetime imaging \cite{Lee.2001}, catalytic activity of fluctuating single lipase molecules\cite{Flomenbom.2005} or stress relaxations in soft matter\cite{Song.2023} that takes place over many different time scales.

\subsection{Physics Flavored Neural Network}
To transform the CNN-Transformer architecture into a Physics-Flavored Neural Network, we integrate our biophysical model directly into the training objective. Building on the formulations in \citeauthor{Luber.2026}, we represent the tetanus contraction as a piecewise function utilizing stretched exponentials for the activation and relaxation phases, augmented by a linear fatigue factor.

\subsubsection{Piecewise Contractile Force Model}

The global force response $F(t)$ is modeled spanning the activation, contraction, and relaxation phases:

$$F(t) = 
\begin{cases} 
0 & t < t_0 \\
F_{\text{rise}}(t) \cdot \Phi(t) & t_0 \leq t \leq t_1 \\
F_{\text{decay}}(t) & t > t_1
\end{cases}$$

where $t_0$ is the onset of stimulation and $t_1 = t_0 + \theta_{\text{dur}}$ is the offset of the tetanic stimulus. During the active phase ($t_0 \leq t \leq t_1$) the force generation follows a stretched exponential growth profile:

$$F_{\text{rise}}(t) = A \left( 1 - \exp\left[ -\left( \frac{t - t_0}{\tau_r} \right)^{\beta_r} \right] \right)$$

To account for force attenuation during sustained stimulation, we apply a multiplicative fatigue factor $\Phi(t)$. To take the biological smoothness into account we model that as a smoothed ReLU with one parameter marks the onset of the fatigue and the other one its slope.

The raw fatigue function $\Phi(t)$ is modeled as a shifted ReLU function convolved by a gaussian kernel to ensure smoothness:

$$\Phi(t) = \alpha ReLU(t-t_{fat})*\phi_{\sigma}$$

with $t_{\text{fat}} = t_0 + t_{\Delta}$ representing the timepoint when the fatigue starts, $t_\Delta$ representive the time delta in respect to the start of the stimulation and $\phi_{\sigma}$  as the probability density function of a gaussian distribution with standard deviation $\sigma$. 
Following the cessation of the stimulus the decay phase ($t > t_1$) spans over the force reached at the end of the stimulus, $A_c = F_{\text{rise}}(t_1) \cdot \Phi(t_1)$, to the baseline $C$ that is reached after full relaxation of the force:

$$F_{\text{decay}}(t) = C + (A_c - C) \exp\left[ -\left( \frac{t - t_1}{\tau_d} \right)^{\beta_d} \right]$$

\subsubsection{Parameter Relaxation and Hybrid Loss}
While several parameters of the physics model are restricted to a certain interval (e.g. $\beta_{r}\in \left[0,1 \right]$, $\alpha \geq 0$ or $t_1 \geq t_0$), the parameter space of the neural network are natively unconstrained. To convert the model parameters to their physically valid domain, we employ sigmoid and softmax transformations, effectively relaxing the parameter space into a differentiable, unconstrained optimization manifold. The detailed list of parameter transformation can be found in Table \ref{tab:parameter_transformations}

\begin{table}[h]
\centering
\caption{Parameter Transformations}
\label{tab:parameter_transformations}
\begin{tabular}{lll}
\hline
\textbf{Physical Parameter} & \textbf{Transformation Equation} & \textbf{Function/Constraint} \\ \hline
$A$             & $A = A_{raw}$                                      & Unconstrained \\
$\tau_r$        & $\tau_r = \text{softplus}(\tau_{r, raw})$          & Positivity \\
$\tau_d$        & $\tau_d = \text{softplus}(\tau_{d, raw})$          & Positivity \\
$\beta_r$       & $\beta_r = \sigma(\beta_{r, raw})$                 & Range $(0, 1)$ \\
$\beta_d$       & $\beta_d = \sigma(\beta_{d, raw})$                 & Range $(0, 1)$ \\
$t_0$           & $t_0 = t_{0, raw}$                                 & Unconstrained \\
$dur$           & $dur = \text{softplus}(\theta_{dur})$              & Positivity \\
$t_1$           & $t_1 = t_0 + dur$                                  & Derived \\
$C$             & $C = C_{raw}$                                      & Unconstrained \\
$t_{\delta}$    & $t_{\delta} = \sigma(t_{\delta, raw}) \cdot 0.8 \cdot dur$ & Scaled Sigmoid \\
$\alpha$         & $\alpha = \text{softplus}(\alpha_{raw})$             & Positivity \\
$\sigma_{smooth}$ & $\sigma_{smooth} = \text{softplus}(\sigma_{raw}) + 10^{-6}$ & Positivity + Epsilon \\ \hline
\end{tabular}
\end{table}

The network is optimized using a hybrid loss function $\mathcal{L}$ which balances supervised parameter regression with physics-based curve reconstruction:

\begin{equation}
    \mathcal{L} = MSE(p_{synth}, \hat{p}_{synth}) + MSE(y_{synth}, F(\hat{p}_{synth})) + MSE(y_{real}, \hat{p}_{real})
\end{equation}

by incorporating the $F(\hat{p})$ reconstruction term, we impose a strong inductive bias on the latent space. The physics loss acts as a rigid prior for interpretability, forcing the model to align its internal tokens with the specific kinetic constants of the stretched exponential. This ensures that the learned parameters are not only accurate but also biologically interpretable, allowing the model to filter out non-physical, high-frequency noise while capturing the fundamental unimodal nature of muscle activation

\subsection{CNN-Transformer Hybrid Architecture}

\begin{figure}
    \centering
    \includegraphics[width=0.99\linewidth]{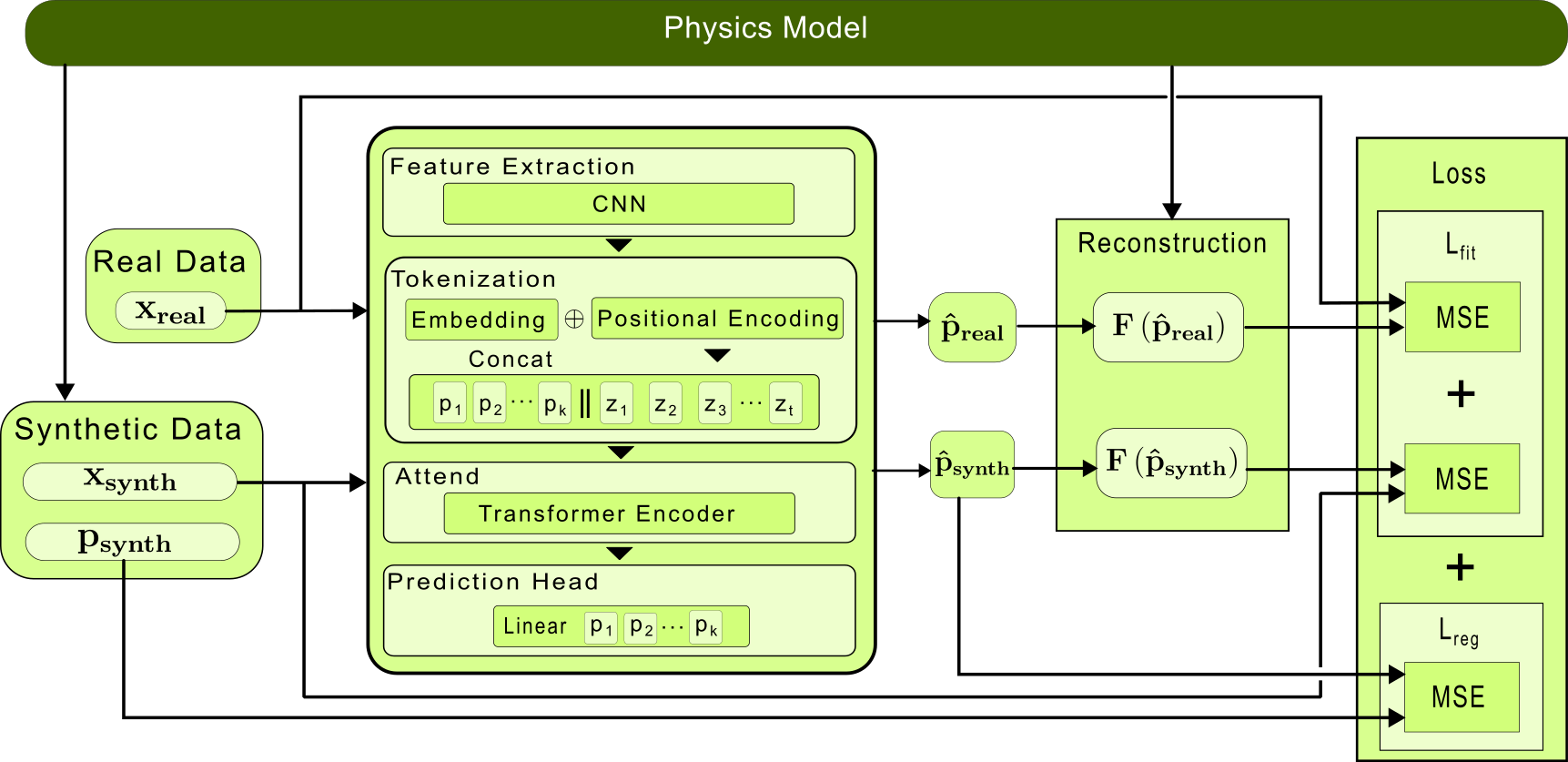}
    \caption{Overview of the physics-flavored CNN-Transformer Architecture. The model itself consists of a few CNN-Layers for the feature extraction followed by an embedding infered with sinusodal positional encoding.  Inspired by the ViT architecture, tokes for the parameter estimation are prepended to the embedding. A transformer encoder with multi-head attention and a linear prediction head is then used to predict the target parameters. The underlying physics model is used once to generate synthetic data with known ground-truth parameters and second as a physics based reconstruction loss. The model is trained on synthetic data by regression model predictions to known-parameter-values and by evaluating the reconstruction error. Unlabeld traing data can be injected into the training procedure in an unsuperwised manner by using the reconstruction loss.}
    \label{fig:ModelArchitecture}
\end{figure}
To learn the kinetic parameters from muscle contraction time-series, we developed a hybrid architecture that combines the local feature extraction of Convolutional Neural Networks (CNNs) with the global context-awareness of Transformers as shown in Figure \ref{fig:ModelArchitecture}. The raw force-time signal is first processed through a series of 1D-convolutional layers. We utilized a respectively large kernel size ($k=15$) to facilitate effective smoothing and to capture the temporal features of the contraction onset and decay. Furthermore, by employing replicate padding instead of zero-padding we mitigate boundary artifacts which is an essential consideration for biological signals where the baseline at $t=0$ and the final relaxation state carry critical physical information. Crucially, batch normalization was omitted in the CNN-layers to prevent the stochastic rescaling of the force signal, ensuring that the model maintains the absolute physical magnitude of the force-time-profiles. 

Following convolutional feature extraction, the signal is projected into a latent space of dimension $d_{model}$ and augmented with sinusoidal positional encodings \cite{Vaswani.2017}. To perform parameterization, we adopted a Vision Transformer (ViT)-inspired approach \cite{Dosovitskiy.2020}, but instead of using a single global classification token as it is common in ViT, we prepended a set of $N$ learnable parameter tokens to latent space, where $N$ corresponds to the number of parameters (N=10) in our physics model. The Transformer Encoder utilizes a multi-head self-attention mechanism to allow these parameter tokens to query the entire signal. This architecture provides two distinct advantages: CNN-Transformers natively handle recordings of different durations without requiring cropping or interpolation of the input. 
Additionally, the attention mechanism of the transformer allows the network to dynamically weight different temporal regions. For example, focusing on the initial onset of the contraction for rise-time parameters, while attending to the long-term plateau for fatigue-related constants.

\subsection{Synthetic Data Generation}
Given the scarcity of experimental data, which is commonly faced in biological experiments, we employed a simulation-to-reality training paradigm. We utilized our stretched exponential physical model to generate an extensive library of 300.000 synthetic contraction curves. By varying the model parameters ($\tau,\beta,\sigma, A_{max}, etc.$) across physiologically relevant ranges and adding randomly scaled and sampled gaussian noise, we produced a diverse set of ground truth labeled data, that was used as the foundation of the training procedure.

\subsection{Training Procedure}
For training the model we implemented a hybrid, semi-supervised training protocol where the optimization objective is a weighted combination of supervised regression and reconstruction on synthetic data and unsupervised physical reconstruction on real-world measurements.

For this we implement an training paradigm that interleaves synthetic and real-world datasets. For each training iteration, the network is primarily exposed to a synthetic data batch $B_{synth}$ where the ground-truth physical parameters $\mathbf{p}_{GT}$ are explicitly known. Simultaneously, to bridge the domain gap, a batch of unlabeled experimental data $B_{real}$ is injected into the training loop with a stochastic probability $p_{real}\in \left [ 0,1 \right]$, where the optimal selection probability was determined via hyperparameter optimization.

This stochastic interleaving compels the model to generalize the deterministic biophysical rules learned from the idealized synthetic library and robustly project them onto the noisy, stochastic environment of empirical \emph{in vitro} recordings.

Reflecting this dual-source data stream, the model is optimized using a composite loss function $L_{total}$ that balances parameter-space accuracy and time-series reconstruction fidelity. 

\begin{equation}
\mathcal{L}_{\text{total}} = \gamma \cdot \mathcal{L}_{\text{fit}}(\mathcal{B}_{\text{syn}}) + (1 - \gamma) \cdot \mathcal{L}_{\text{param}}(\mathcal{B}_{\text{syn}}) + \delta_{\text{real}} \cdot \lambda \cdot \mathcal{L}_{\text{fit}}(\mathcal{B}_{\text{real}})
\end{equation}

where $\gamma \in \left[0,1\right]$ balances the synthetic training objectives, $\lambda >0$ is a prefactor weighting the real-world domain adaptation, and $\delta_{\text{real}}$ is drawn from: $ \delta_{\text{real}} \sim \text{Bernoulli}(P_
{\text{real}})$

The parameter loss ($\mathcal{L}_{param}$) uses the mean squared error (MSE) loss calculated between the predicted and ground-truth parameters for synthetic data, ensuring the model converges toward the correct biophysical constants. The physics-fit loss ($\mathcal{L}_{fit}$) also used an MSE loss, but between the input force-time signal and the curve reconstructed by passing the predicted parameters through the physics flavored model $F(\hat{p})$.
Additionally to training parameters like  $\lambda$, $\gamma$ and $p_{real}$  architectural parameters were selected by extensive hyperparameter optimization through the Optuna framework \cite{Akiba.2019}. We tuned our CNN-Transformer specifically for its ability to achieve a low reconstruction error on the real-world measurments of the validation set. $$\text{Objective} = \frac{1}{N} \sum_{i=1}^{N} \text{MSE}(y_{real, i}, F(\hat{p}_{real, i}))$$

\section{Results}\label{sec2}

\subsection{Synthetic Data is a strong basis for predicting real measurements}
\begin{figure}
    \centering
    \includegraphics[width=0.99\linewidth]{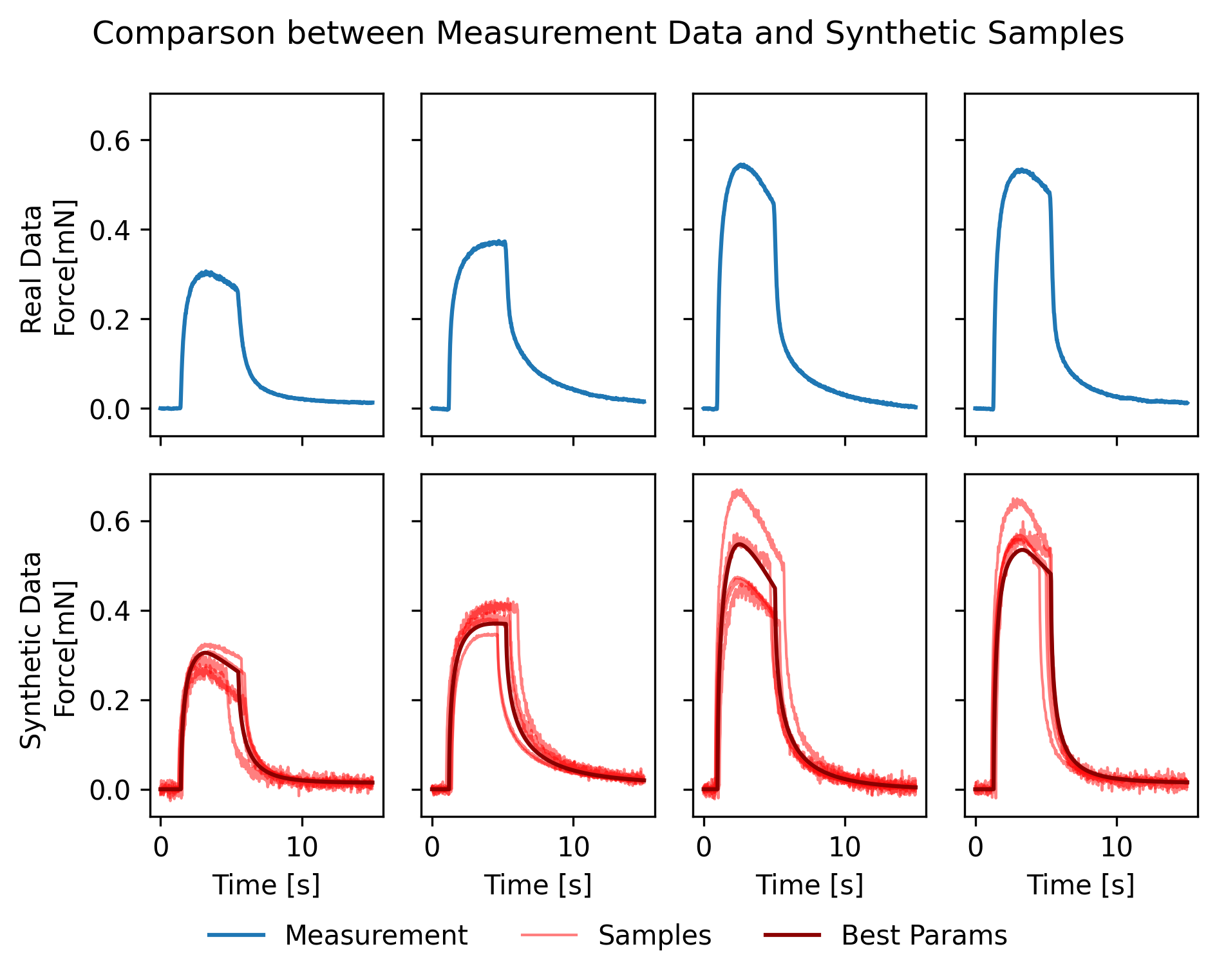}
    \caption{The upper row displays experimental force-time profiles (blue), illustrating a wide spectrum of contractile phenotypes. The contraction curves do not only show differences in terms of their absolute peak force, but also in terms of fatiguing behavior or the timescale of contraction/relaxation. The data was obtained from \citeauthor{Luber.2026} and the differences in contraction dynamics where induces by inhibiting different molecular mechanism via  pharmacological intervention.\cite{Luber.2026}. For direct comparison, the synthetic cohorts are centered around the respective experimental curves: the optimal parameter fit is highlighted in dark red, while surrounding lines (light red) represent alternative parameter permutations to illustrate the generative variance. The synthetic profiles are visually near-indistinguishable from the real measurements, capturing the non-linear activation rise, plateau stability, and relaxation kinetics. The full synthetic dataset spans randomly generated parametrizations designed to entirely cover the empirical distribution of the experimental curves.}
    \label{fig:synth_vs_real}
\end{figure}

The cornerstone of this physics-flavored approach is the ability of the underlying mathematical model to represent the true biological signal.  We previously showed that the proposed model is able to capture a broad set of kinetics profiles of ESM contraction curves \cite{Luber.2026}. As demonstrated in Figure \ref{fig:synth_vs_real}, we can show now that synthetic data generated with this physics-flavoured model (dark red) yields artificial contraction curves that are visually indistinguishable from real experimental data (blue). The four representive samples give an impression about the range of contractile phenotypes contained in the experimental data. It includes characteristic contraction curves with force-plateaus, different degrees of fatiguing behaviour, but also  pharmacologically inhibited samples with drastically lowered peak force and heavily altered contraction/relaxation velocities. 

\begin{figure}
    \centering
    \includegraphics[width=0.99\linewidth]{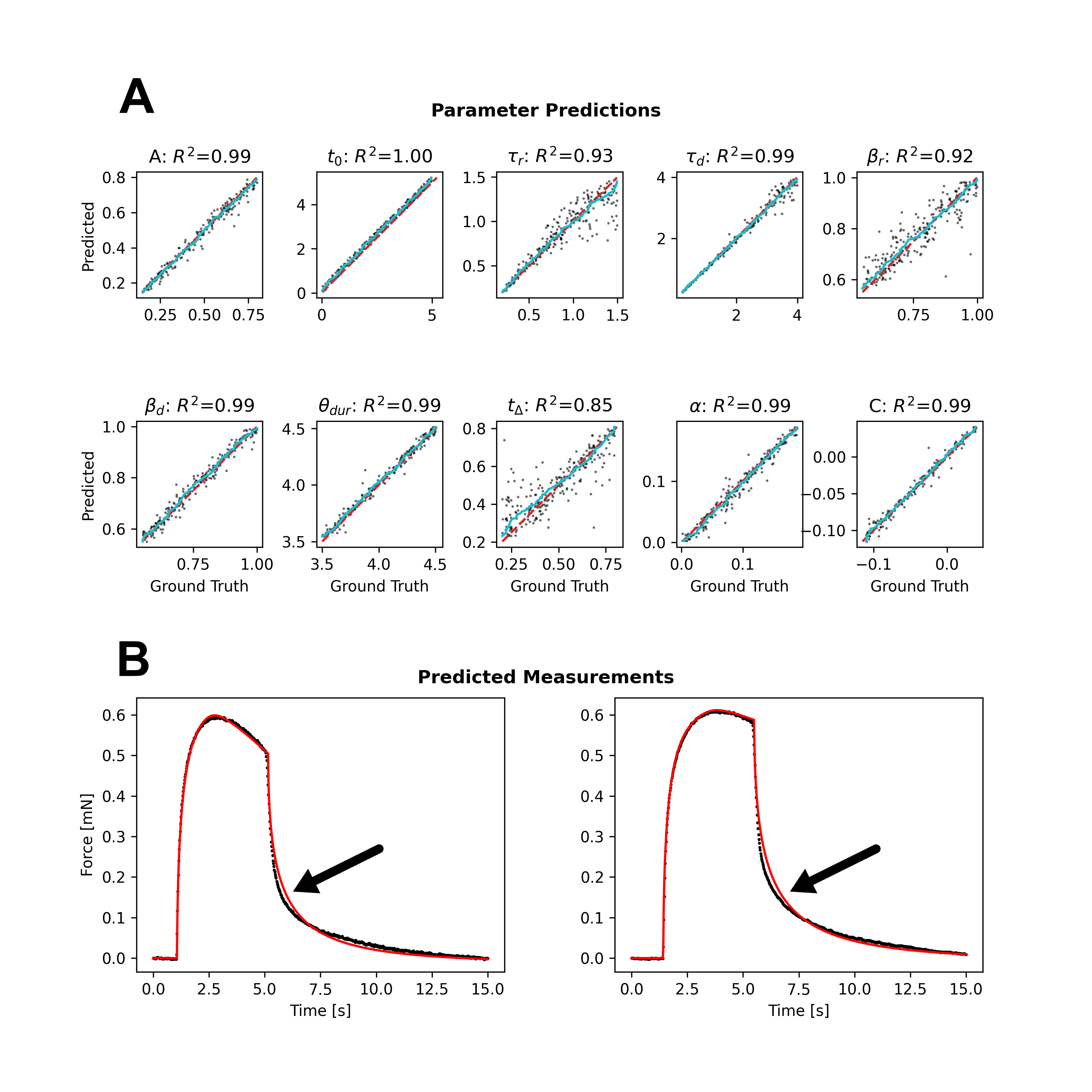}
    \caption{  Quantitative and qualitative evaluation of the CNN-Transformer network trained exclusively on synthetic data. 
    \textbf{A}: Scatterplot comparing network-predicted physical parameters against synthetic ground truth on an independent hold-out test set. Across all ten parameters of the stretched exponential model, the network achieves high to near-perfect correlation, with coefficient of determination ($R^2$) values ranging from $0.85$ to $1.00$. This correlation is highlighted by the empirical Quantile-Quantile-line (cyan) against the identity line (red dashed).
    \textbf{B}: Qualitative zero-shot evaluation of the synthetically trained model applied to empirical \emph{in vitro} recordings (black dots). While the model accurately reconstructs the overall contraction dynamics, the upstroke, and also the peak force, a slight but consistent structural deviation (red line) is visible during the decay regime.} 
    \label{fig:enter-label}
\end{figure}

The synthetic data (red) generated by the physics model matches the appearance of the real experimental data closely and is visually difficult to distinguish from the real measurement data. Both types of data capture the characteristic non-linear activation rise, the sustained (and potentially decaying) plateau, and the long-tailed relaxation kinetics. 
This high degree of morphological overlap confirms that the stretched exponential model provides a sufficient and robust descriptor of the complex, multi-modal dynamics inherent in engineered skeletal muscle contractions.

To first verify that the model successfully learns the inverse mapping of the physics-flavored model and can accurately predict model parameters from contraction curves, we evaluated the coefficient of determination (R²) and quantile–quantile plots for each parameter (Figure \ref{fig:enter-label}). To this end the model is trained exclusively on synthetically generated data where the true parameters are known. Across all parameters the predictions are remarkably close to the ground truth values. While the peak-force $A$, the fatigue-slope $\alpha$, the on-set and duration of the contraction as well as the decay parameters $\tau_d,\beta_d $ reach $R^2$ coefficients close to a perfect value of 1, the rise parameters $\tau_r,\beta_r$ score slightly worse. This is probably due to the shorter duration of the signal for the rise time compared to the relaxation phase and fatiguing behavior inferring with the rise. The largest, yet still acceptable, deviations between the predicted and ground-truth values were observed for the onset of fatigue $t_\Delta$. This is expected because i.e. very small onsets or very large values are hard to predict. Small values might be hard to predict, as in these cases the curves have not reached their plateau yet, whereas for high values the remaining time before the relaxation might be too short to identify the offset correctly for noisy curves. Conversely, large $t_\Delta$ values correspond to short fatigue phases, providing fewer data points and less information for the model to distinguish it from the decay onset

To assess the predictive performance on empirical data, we evaluated the model on unlabeled real-world recordings of the validation set.
Despite having been trained exclusively on synthetic distributions without prior exposure to biological signals, the model demonstrates a remarkably robust zero-shot baseline performance. 
It successfully generalizes the abstract mathematical principles learned from the synthetic library to the noisy, stochastic environment of \emph{in-vitro} assays.
While the overall contraction morphology is captured with high fidelity, subtle but noticeable deviations are occasionally observed in the decay parameters, as illustrated in Figure\ref{fig:enter-label} B.

Aside from these small deviations, we conclude that the synthetic data distribution is sufficiently dense and physically accurate to provide the foundational information required for learning the fundamental shape of contraction curves. Nevertheless, it also highlights nuanced differences between the idealized physics-flavored model and noisy \emph{in-vitro} measurements. This underscores the necessity for the subsequent semi-supervised integration of empirical measurements into the training procedure to sensitize the model to information not yet captured by the underlying equations.

\subsection{Injecting unsupervised real measurement data into the training procedure improves the prediction}

\begin{figure}
    \centering
    \includegraphics[width=0.99\linewidth]{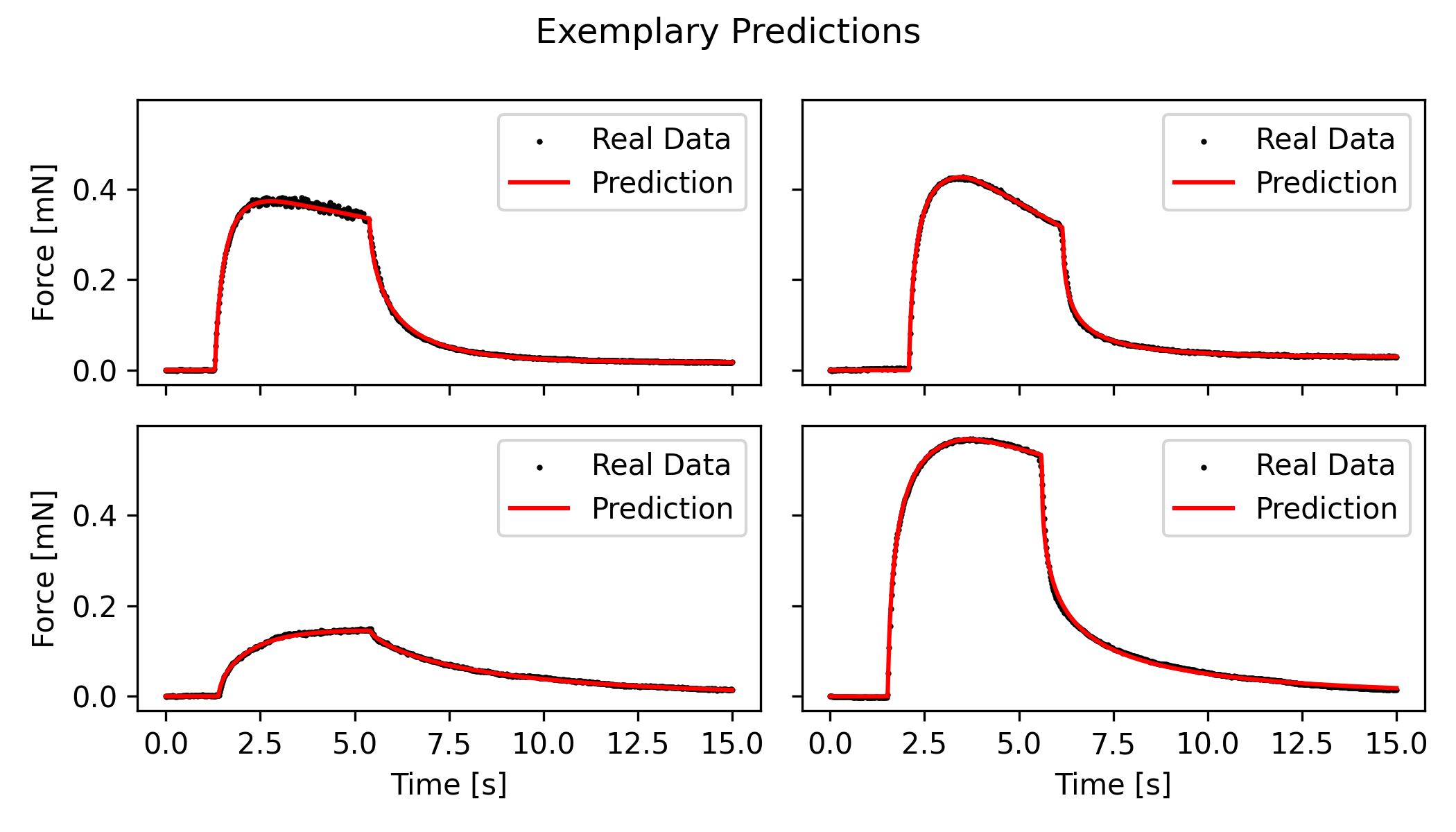}
    \caption{Qualitative zero-shot evaluation of the semi-supervised model, trained on synthetic data and real \emph{in vitro} recordings, using a held-out test dataset. The real \emph{in-vitro} recordings are shown in black, the contraction curve reconstructed by the predicted parameters are shown in red. Across different contractile phenotypes the fit matches the empirical data almost perfectly. In comparison to the synthetic model as shown in Figure\ref{fig:enter-label} B the prediction of the relaxation phase improved noticeably.}
    \label{fig:real_pred}
\end{figure}

While the synthetic baseline training already demonstrates a strong physical intuition, we hypothesized that the model's performance could further improved by adapting to the subtle nuances of experimental situations. This includes sensor noise and other sources of non-idealized biological variance that are difficult to model synthetically. By injecting unlabeled real-world measurement data into the training loop and adding physics-flavored reconstruction loss to the training objective, we observed a noticeable enhancement in fit fidelity for real measurements. We find that the slight misprediction in the relaxation phase vanished and the model is now able to capture the contraction dynamics accurately over a broad set of different contractile phenotypes (Figure \ref{fig:real_pred}).
This improvement demonstrates a key advantage of our framework: the model does not require human-labeled ground truth for predicting real empirical data accurately. Instead, it uses the mathematical constraints of the stretched exponential to label the real data itself. By minimizing the discrepancy between the raw biological signal and the physically-valid reconstruction, the network undergoes a process of unsupervised self-alignment, effectively bridging the domain gap between idealized math and complex, 3D engineered tissue.

\subsection{Temporal attribution derived by integrated gradients show co-localization with the effective parameter regimes}

To verify that our CNN-Transformer architecture had internalized the biophysical principles of the stretched exponential model rather than merely memorizing signal shapes we employed Integrated Gradients (IG) \cite{Sundararajan.2017,Kokhlikyan.2020}. This attribution method maps the network's internal decision-making process by identifying which temporal regions of the force-time signal are most influential for specific parameter predictions. Regions in the signal with absolute high values for the feature attribution scores are considered to have an high influence on the prediction for the respective parameters.

\begin{figure}
    \centering
    \includegraphics[width=0.99\linewidth]{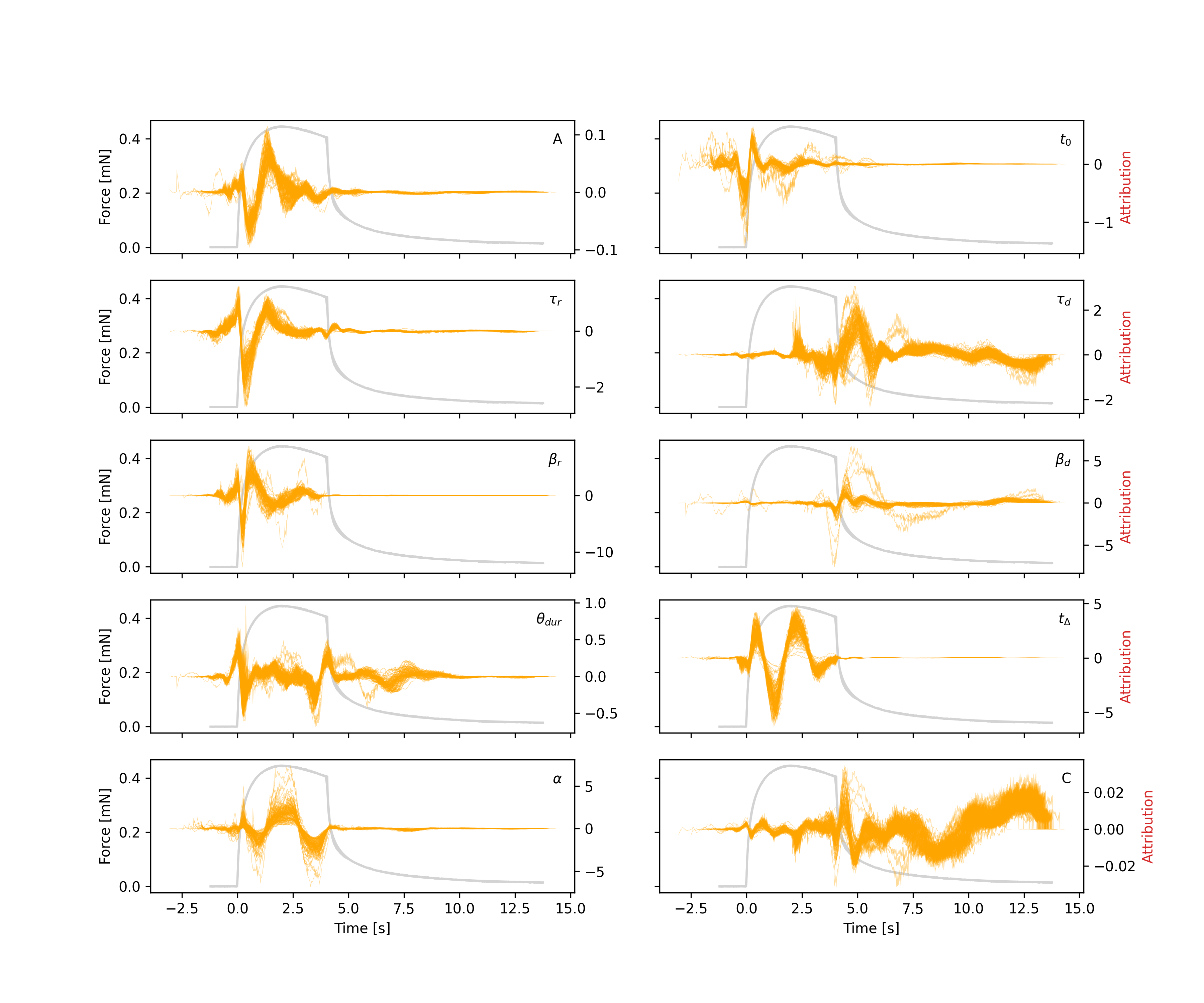}
    \caption{Attribution scores for the different parameter predictions. The plots illustrate the temporal regions within the force-time profiles that most heavily influence the network's prediction for each individual physical parameter. Feature attribution scores (orange curves) were computed for all samples in the independent hold-out test set and temporally aligned relative to the contraction onset ($t_0 = 0\,\text{s}$). For physiological context, a representative contraction curve is superimposed in gray (left axis: Force $[\text{mN}]$; right axis: Attribution score). Regions exhibiting high absolute attribution values signify a greater impact on the respective parameter estimation. The distinct, parameter-specific colocalization of attribution peaks within corresponding physiological regimes e.g. such as upstroke parameters ($\tau_r, \beta_r$) clustering at the activation onset and decay parameters ($\tau_d, \beta_d$) targeting the relaxation phase, indicates that the model successfully attend to the relevant regime of the contraction curve for predicting the respective parameter values.}
    \label{fig:IG}
\end{figure}

As shown in Figure \ref{fig:IG} we derived the temporal attribution maps throughout all real-world measurements in the test set.  To ensure comparability the attribution scores (orange) were aligned relative to the contraction onset ($t_0 = 0\,\text{s}$) and for contextualization an exemplary contraction curve was added. 
Strikingly, the attribution profiles demonstrate a high degree of temporal co-localization with the physiological regimes governed by their respective parameters. Parameters responsible for modeling the initial contractile phase (such as $\tau_r$ and $\beta_r$) show sharp attribution peaks concentrated heavily along the upstroke of the contraction curve and conversely, the attribution scores for the decay parameters ($\tau_d$ and $\beta_d$) span across the long tail of the relaxation phase.
Most notably, the attribution profile for the contraction duration parameter ($\theta_{dur}$) exhibits a distinct bimodal signature, spiking sharply at both the onset and the offset of the contraction. This indicates that the network correctly identifies the temporal boundaries of the contraction phase and uses that information for predicting the duration of the contraction $\theta_{dur}$. Similarly, the offset parameter $C$ is correctly and prominently focused on the terminal regime of the relaxation phase.

\subsection{Model predictions on real measurement data surpasses L-BFGS in terms of accuracy}\label{sec:benchmarking}

\begin{figure}
    \centering
    \includegraphics[width=0.99\linewidth]{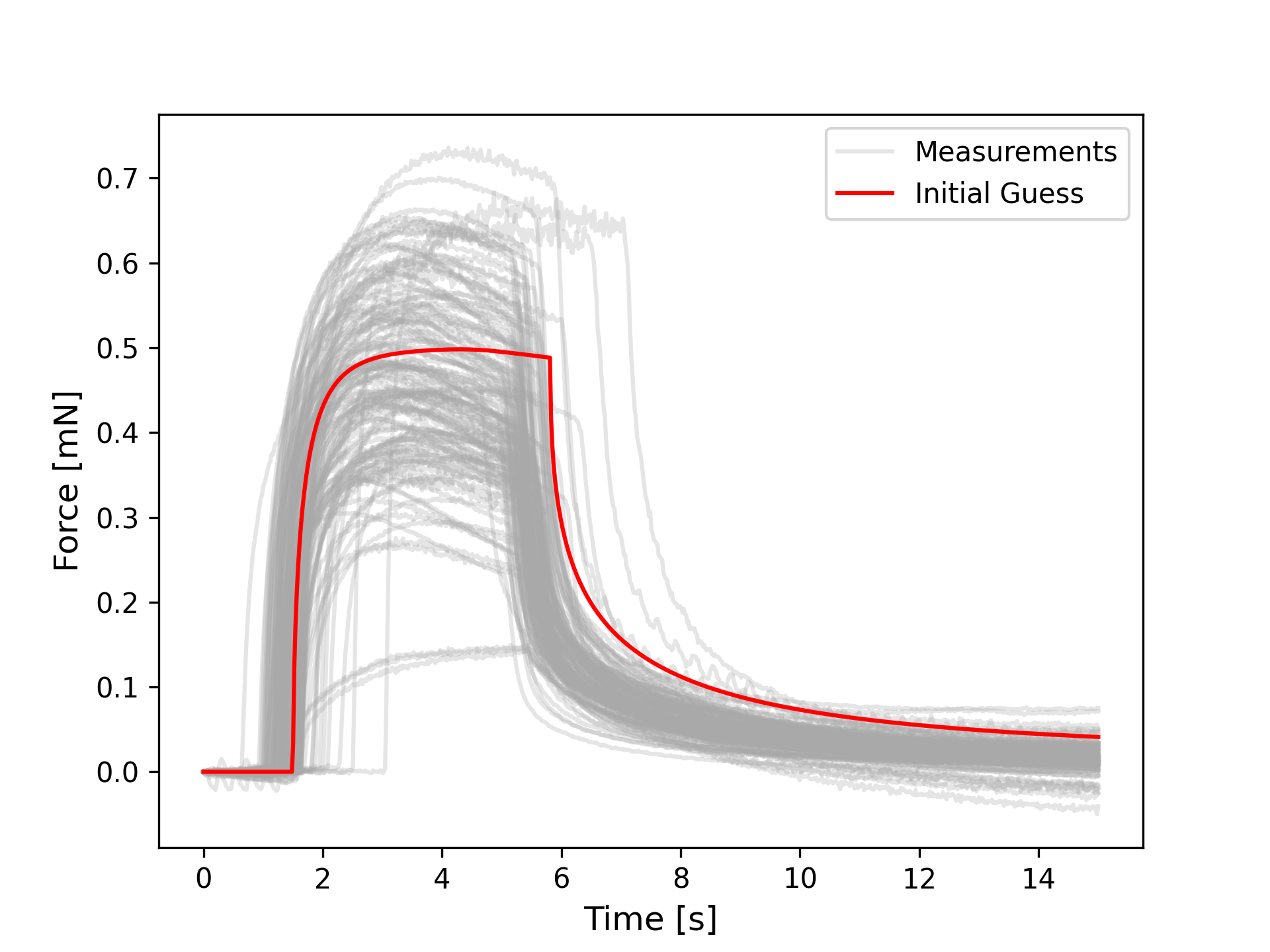}
    \caption{Visualization of the initial parametrization for the L-BFGS-based parameter estimation. The initialization (red) was chosen to represent an "average" contraction curve. This initialization was then used as a starting point to fit the experimentally derived measurement curves (gray) with the L-BFGS algorithm. }
    \label{fig:initial_guess}
\end{figure}

To evaluate the practical utility of our framework we benchmarked the CNN-Transformer’s predictions against a standard L-BFGS (Limited-memory Broyden-Fletcher-Goldfarb-Shanno) \cite{Liu.1989,Schmidt.2005} iterative solver and additionally assessed the effect of the hybrid training strategy quantitatively based on the mean squared error (MSE) score on both the validation set.

In order to derive a benchmark for the L-BFGS we initialized the fitting procedure with a reasonable parameter guess that represents an averaged contraction curve (Figure \ref{fig:initial_guess}) and then optimized until convergence. 

As shown in Table \ref{tab:results}, the neural network's direct parameter predictions on the validation set resulted in lower mean squared error compared to the baseline L-BFGS optimization. While the classical optimizer apparently became trapped in local minima, leading to non-physical fits that failed to capture the contraction curves accurately, the physics-flavored model was able to identify the correct physical regime directly.
This suggests that the model has internalized the distribution of possible contraction dynamics, allowing it to provide a high-quality physical prior even for complex, non-standard contraction profiles. 
Again, the injection of unlabeled-measurement data into the training procedure noticeably improved the prediction accuracy by a factor of 2.5 in comparison to the model that was exclusively trained on synthetic data.

To check whether the fit could be further improved by fine-tuning the contraction curves, we applied the L-BFGS curve fitting on-top of the model predictions. However, in this cases only a marginal improvement could be achieved, which indicates that the model score close to the optimal predictions.

This concludes that our architecture mitigates the need for manual parameter tuning or "average-guess" heuristics, providing a fully automated, high-fidelity pipeline that is both faster and more accurate than traditional iterative methods alone.

\begin{table}[ht]
\centering
\caption{Mean Squared Error (MSE) and standard deviation (SD) on the validation dataset across baseline L-BFGS optimization (LBFGS), our model trained on synthetic data only (Our (Synth)), our model with the hybrid training with synthetic and real data (Ours (Hybrid)), and hybrid training with post-hoc L-BFGS fine-tuning (Ours (Hybrid) + LBFGS)}
\label{tab:results}

\begin{tabular}{lrr}
\hline
 & MSE & SD \\
\hline
LBFGS & 0.000417 & 0.004070 \\
Ours (Synth) & 0.000044 & 0.000040 \\
Ours (Hybrid) & 0.000017 & 0.000012 \\
Ours (Hybrid) + LBFGS & 0.000015 & 0.000011 \\
\hline
\end{tabular}

\end{table}

\subsection{Model generalizes to additional cell lines}
While the experimental data used for training capture a broad range of contractile phenotypes through pharmacological modulation, they nevertheless originate from a single cell line (LHCN-M2) derived from the \emph{pectoralis major}  chest muscle of a 41-year-old male donor. 

To evaluate the generalization of our physics-flavored approach across varying physiological and pathological conditions, we utilized the three secondary cell lines that represent healthy baseline (AB1167), dystrophic (KM571), and aging (KM670) muscle paradigms, respectively.

\begin{figure}
    \centering
    \includegraphics[width=0.99\linewidth]{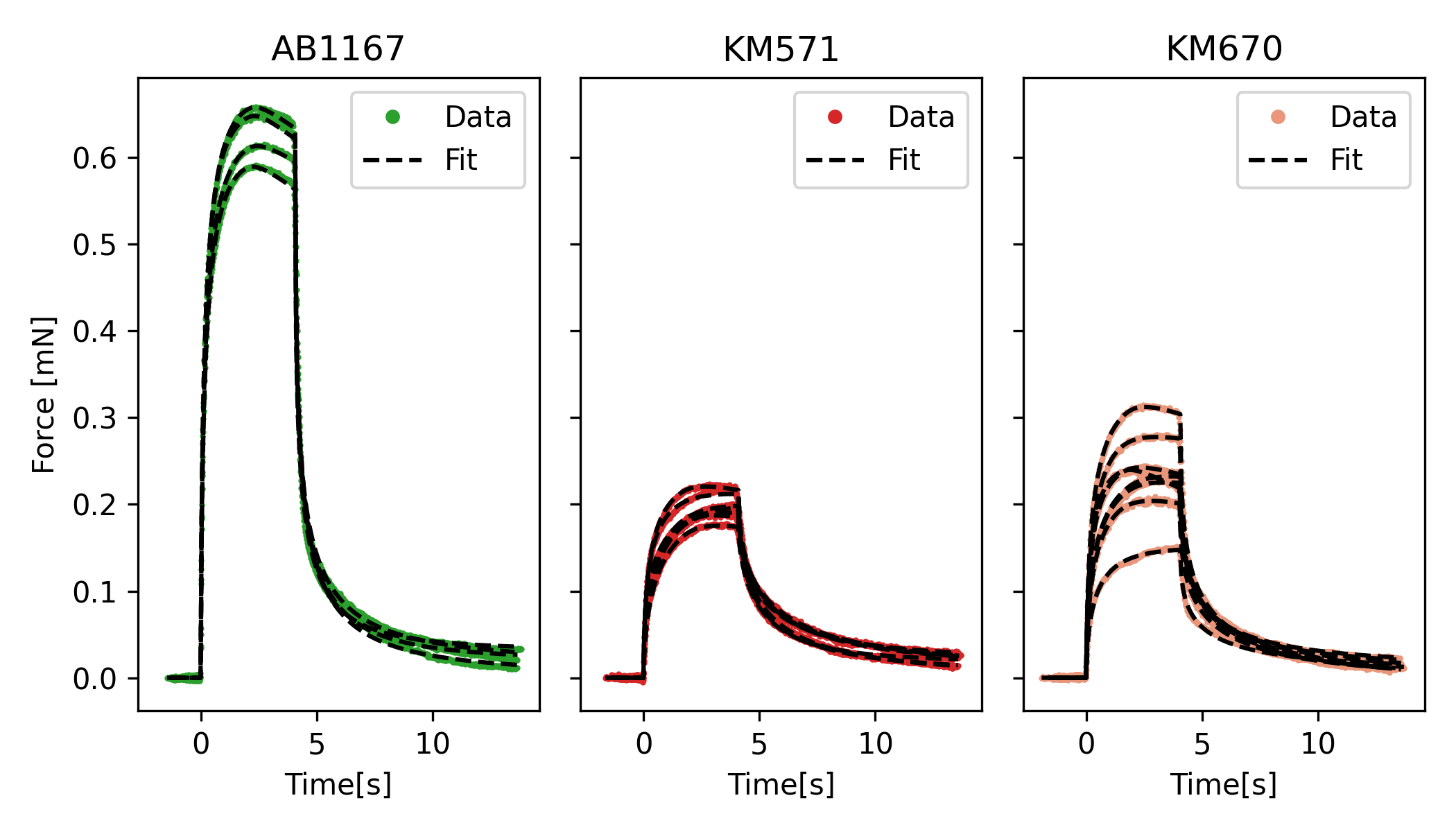}
    \caption{Visualization of the fit derived by the model (black) and the measurement data (colored) for the three different test cell lines. Despite differences in anatomical muscle origin, age and disease status the model is able to fit the respective contractile phenotype accurately.}
    \label{fig:cell_lines}
\end{figure}

As shown in Figure \ref{fig:cell_lines}, the model generalizes remarkably well across all three lines, despite profound differences in donor age, disease state, and anatomical context compared to the primary training data. Consistent with the benchmarks established in Section \ref{sec:benchmarking}, our network outperforms the classical L-BFGS fitting strategy (see Table \ref{tab:table2}). However, initializing a subsequent L-BFGS refinement step on top of the network’s parameter predictions yielded minor improvements in fit fidelity.

Aside from accurate predictions, the parameterization achieved through our physics-based model enables comprehensive insights into the contraction-relaxation cycle beyond simple peak force measurements by inspecting the kinetic parameters from the latent space. Our analysis reveals that the investigated cell lines differ not only in absolute force generation but also show prominent differences in their contraction and relaxation behavior.

The peak force of the healthy wild-type tissue (AB1167) is notably higher than that of both the pathological (KM571) and aged (KM670) samples. Concurrently, the overall velocity of both contraction and relaxation is significantly decayed in the DMD and aged tissues. This finding aligns with other literature reporting reduced peak force and prolonged relaxation kinetics in dystrophic and sarcopenic muscle.\cite{Mestre.2021,Larsson.2019,Korhonen.2006, NicolasMetral.2001,Goldstein.2010,Nesmith.2016}

While the heterogeneity parameters governing the contraction upstroke differ only slightly across all samples, the parameter $\beta_d$ reveals substantially greater kinetic heterogeneity in the healthy control ($\beta_d = 1$ corresponds to a single-exponential decay). One possible explanation for this lies in the different timescales involved in muscle relaxation. While the cytosolic calcium concentration in healthy muscle declines more rapidly than force relaxation \cite{Rausch.2020}, it is well established that impaired calcium handling in DMD muscle delays relaxation by slowing the re-uptake of calcium into the sarcoplasmic reticulum following contraction\cite{Riddell.2023,Burr.2015,Mareedu.2021}. This slowing could shift the usually fast timescales of the calcium handling closer to the slower timescale of the passive relaxation of the ECM, consequently leading to a more homogeneous decay. While the current study is limited to only three cell lines a larger cohort is of course required to validate this hypothesis. 

\begin{figure}
    \centering
    \includegraphics[width=0.9\linewidth]{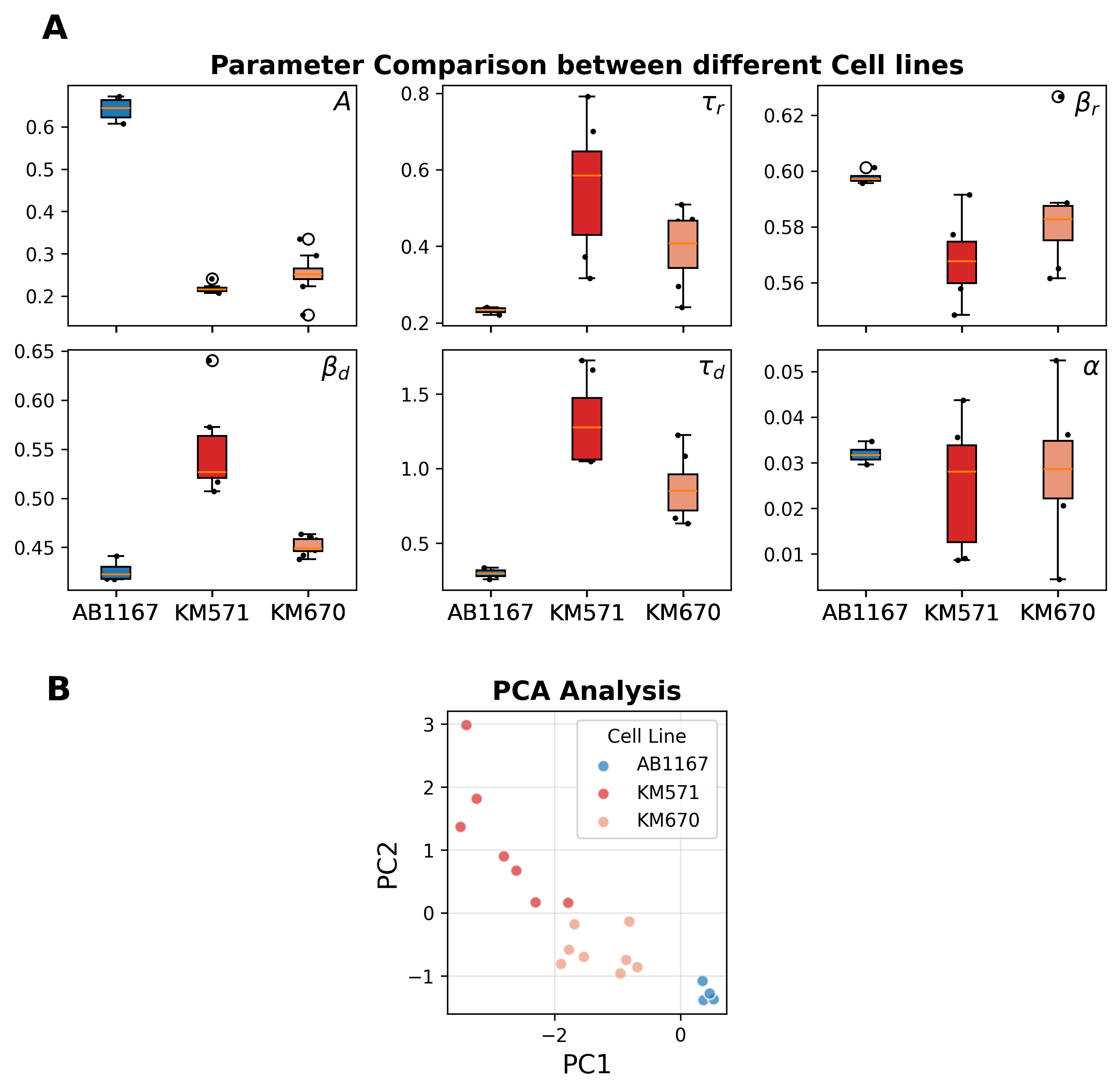}
    \caption{\textbf{A)} Comparison of the predicted contractile parameters for different cell lines. The healthy cell line AB1167 shows a much higher peak force and faster contraction/relaxation dynamics compared to the diseased (K571) or aged (KM670) counterparts. Whereas the heterogenity parameter for the contraction only shows mild deviation between the cell lines, the $\beta_d$ paramters for the DMD cell line is substatially higher, indicating a relaxation closer to a mono exponential behaviour. $\alpha$ values close to zero for all cell lines indicate no fatiguing behaviour. \textbf{B)} Principal Component Analysis of the predicted parameters. A clear separation of the different cell lines can already be seen by plotting the first two principal components against each other.}
    \label{fig:PCAPlot}
\end{figure}

The Principal Component Analysis (PCA) in  Figure \ref{fig:PCAPlot} computed from the predicted parameter vectors confirms these distinct biophysical profiles, as it shows clusters for each cell line. This clustering demonstrates that the physics-flavored parameter space effectively extracts a physiological fingerprint capable of separating aging and disease phenotype.

Together, the generalization across diverse muscle tissues and the insights gained by the parameter-based profiling of the contraction dynamics highlight the potential of the PFNN framework for automated, high-content screening in muscle research.

\begin{table}[ht]
\centering
\caption{Model performance on contraction data obtained from additional cell lines. Values are reported as mean $\pm$ standard deviation.}
\label{tab:table2}

\begin{tabular}{llll}
\hline
 & LBFGS & Ours (Hybrid) & Ours (Hybrid) + LBFGS \\
Cell line &  &  &  \\
\hline
AB1167 & 0.000064 ($\pm$0.000027) & 0.000052 ($\pm$0.000010) & 0.000044 ($\pm$0.000010) \\
KM571  & 0.000013 ($\pm$0.000014) & 0.000007 ($\pm$0.000002) & 0.000004 ($\pm$0.000001) \\
KM670  & 0.000040 ($\pm$0.000100) & 0.000010 ($\pm$0.000003) & 0.000007 ($\pm$0.000002) \\
\hline
\end{tabular}

\end{table}

\section{Discussion}

The primary contribution of this work is the development of an applicable framework to accurately and autonomously parametrize tetanus-contraction curves by employing a physics-flavoured CNN-Transformer network. The beauty of the proposed architecture lies in the strategic blending of its components. CNN layers provide the initial pre-processing of the model, leveraging their well-known capacity for local pattern recognition and noise filtering within time-series data. While it has been shown that CNNs are surprisingly good in learning absolute positions \cite{Islam.2020}, attention-based transformers excel at learning global context\cite{Vaswani.2017}. Nevertheless, it would have been possible to get similar networks with deeper CNN layers followed by an adaptive-pooling \cite{He.2015} to map variable length signals to a fixed set of parameters, or by the integration of recurrent neural networks like GRU \cite{Chung.2014} or LSTM\cite{Vennerd.2021}. However, we believe that the combination of shallow CNNs and transformer encoder layers solves this conceptually more elegant.

By treating each parameter prediction as a distinct ViT-style token, the network effectively creates a dynamically generated knowledge bank throughout the forward pass that queries the entire contraction sequence and that can be used for the prediction of the paramters in the final linear layer. The attention mechanism allows the model to "puzzle together" local features such as the initial rise slope or the onset of fatigue, and write that information into the respective parameter tokens. This modularity promotes selectively attending to the specific temporal windows relevant to each kinetic constant.

A significant hurdle in applying Transformer-based models to biology is the common scarcity of experimental data. Transformers usually require large amounts of data, yet biological datasets are notoriously small and expensive to label \cite{Anwar.2025}. We overcome this by utilizing the stretched exponential model in a dual role:

First, it acts as a generative engine for synthetic pre-training, providing the network with a physics rooted prior and a labeled corpus for heavy-duty optimization. Second, by integrating the model as a differentiable PINN loss, we enable the network to ingest unlabeled experimental data. This hybrid approach allows the model to fine-tune its representations on real-world noise and biological variability without requiring manual ground-truth labels. The resulting pipeline is thus self-reinforcing and scales well as more experimental data is collected. For further labeling of the measurement data the trust-region solvers could be applied on top of the networks prediction to label experimental data in an automatic way.

\section{Limitations}
While the current framework offers a robust solution for scaleable phenotyping, several avenues for optimization remain. Currently, the model utilizes standard self-attention, which carries a quadratic computational complexity ($\mathcal{O}  L^2 $). While manageable for 15-second recordings at 50 FPS, moving toward linear attention mechanisms \cite{Choromanski.2020} or linear state-space models like Mamba \cite{Gu.2023} could enable continuous, real-time live analysis of muscle maturation over longer periods of time or at higher frequencies.

Furthermore, our model currently is trained on a fixed sampling rate of 50 fps, which we relaxed a bit via data augmentation. A more elegant future iteration could replace the current sinusoidal positional encodings with the absolute measurement timestamps ($t_i$). This would allow the model to be inherently agnostic to sampling frequency and unevenly spaced data points, further increasing its utility across different experimental hardware.

\section{Conclusion}

In summary, by embedding the physics of skeletal muscle contraction directly into an attention-based architecture, we have moved beyond traditional, one-dimensional metrics such as recording peak forces, and appraoch a more comprehensive, kinetic phenotyping of engineered tissues. By synergizing a physics-grounded theoretical model with a CNN-Transformer architecture, and employing a hybrid training paradigm that leverages both unsupervised real-world measurements and synthetic data, we have developed a robust, automated pipeline.
This "Physics-Flavored" approach mitigates the bottleneck of sparse biological annotations and scales seamlessly with expanding datasets, providing an automated pipeline for high-throughput kinetic phenotyping across drug perturbations, disease phenotypes, and age-related functional decline.
\newpage

\section*{ACKNOWLEDGMENTS}

TB was supported by the Deutsche Forschungsgemeinschaft (DFG) under project 569108445. This work was supported by the European Union’s Horizon Europe research and innovation programme under grant agreement No. 101247478. 

\section*{AUTHOR CONTRIBUTIONS}

Conceptualization, M.L. ; methodology, M.L.; implementation, M.L., investigation, M.L. and T.B.; writing-original draft, M.L.; writing-review \& editing, M.L. and T.B.; funding acquisition, T.B.; supervision, T.B.

\section*{DECLARATION OF INTERESTS}
The authors declare no competing interests

\section*{DECLARATION OF GENERATIVE AI AND AI-ASSISTED TECHNOLOGIES}

During the preparation of this work the authors used Gemini in order to improve the language, flow, and clarity of the manuscript through proofreading and text formulation. After using this tool/service, the authors reviewed and edited the content as needed and take full responsibility for the content of the published article.

\newpage

\newpage

\bibliography{references}

@article{AfsharBakooshli.2019,
 author = {{Afshar Bakooshli}, Mohsen and Lippmann, Ethan S. and Mulcahy, Ben and Iyer, Nisha and Nguyen, Christine T. and Tung, Kayee and Stewart, Bryan A. and {van den Dorpel}, Hubrecht and Fuehrmann, Tobias and Shoichet, Molly and Bigot, Anne and Pegoraro, Elena and Ahn, Henry and Ginsberg, Howard and Zhen, Mei and Ashton, Randolph Scott and Gilbert, Penney M.},
 year = {2019},
 title = {A 3D culture model of innervated human skeletal muscle enables studies of the adult neuromuscular junction},
 volume = {8},
 journal = {eLife},
 doi = {10.7554/eLife.44530}
}

@article{Ahmadi.2026,
 author = {Ahmadi, Nazanin and Cao, Qianying and Humphrey, Jay D. and Karniadakis, George Em},
 year = {2026},
 title = {Physics-Informed Machine Learning in Biomedical Science and Engineering},
 pages = {309--336},
 volume = {28},
 number = {1},
 journal = {Annual review of biomedical engineering},
 doi = {10.1146/annurev-bioeng-110824-124907}
}

@misc{Akiba.2019,
 author = {Akiba, Takuya and Sano, Shotaro and Yanase, Toshihiko and Ohta, Takeru and Koyama, Masanori},
 date = {2019},
 title = {Optuna: A Next-generation Hyperparameter Optimization Framework},
 publisher = {arXiv},
 doi = {10.48550/ARXIV.1907.10902}
}

@article{Anwar.2025,
 author = {Anwar, Ayman and Khalifa, Yassin and Coyle, James L. and Sejdic, Ervin},
 year = {2025},
 title = {Transformers in biosignal analysis: A review},
 pages = {102697},
 volume = {114},
 issn = {15662535},
 journal = {Information Fusion},
 doi = {10.1016/j.inffus.2024.102697}
}

@article{BerberanSantos.2005,
 author = {Berberan-Santos, M. N. and Bodunov, E. N. and Valeur, B.},
 year = {2005},
 title = {Mathematical functions for the analysis of luminescence decays with underlying distributions 1. Kohlrausch decay function (stretched exponential)},
 pages = {171--182},
 volume = {315},
 number = {1-2},
 issn = {03010104},
 journal = {Chemical Physics},
 doi = {10.1016/j.chemphys.2005.04.006}
}

@article{Burr.2015,
 author = {Burr, A. R. and Molkentin, J. D.},
 year = {2015},
 title = {Genetic evidence in the mouse solidifies the calcium hypothesis of myofiber death in muscular dystrophy},
 pages = {1402--1412},
 volume = {22},
 number = {9},
 journal = {Cell death and differentiation},
 doi = {10.1038/cdd.2015.65}
}

@article{Cai.2021,
 author = {Cai, Shengze and Mao, Zhiping and Wang, Zhicheng and Yin, Minglang and Karniadakis, George Em},
 year = {2021},
 title = {Physics-informed neural networks (PINNs) for fluid mechanics: a review},
 pages = {1727--1738},
 volume = {37},
 number = {12},
 issn = {0567-7718},
 journal = {Acta Mechanica Sinica},
 doi = {10.1007/s10409-021-01148-1}
}

@article{Cheng.2014,
 author = {Cheng, Cindy S. and Davis, Brittany N. J. and Madden, Lauran and Bursac, Nenad and Truskey, George A.},
 year = {2014},
 title = {Physiology and metabolism of tissue-engineered skeletal muscle},
 pages = {1203--1214},
 volume = {239},
 number = {9},
 journal = {Experimental biology and medicine (Maywood, N.J.)},
 doi = {10.1177/1535370214538589}
}

@misc{Choromanski.2020,
 author = {Choromanski, Krzysztof and Likhosherstov, Valerii and Dohan, David and Song, Xingyou and Gane, Andreea and Sarlos, Tamas and Hawkins, Peter and Davis, Jared and Mohiuddin, Afroz and Kaiser, Lukasz and Belanger, David and Colwell, Lucy and Weller, Adrian},
 date = {2020},
 title = {Rethinking Attention with Performers},
 publisher = {arXiv},
 doi = {10.48550/ARXIV.2009.14794}
}

@misc{Chung.2014,
 author = {Chung, Junyoung and Gulcehre, Caglar and Cho, KyungHyun and Bengio, Yoshua},
 date = {2014},
 title = {Empirical Evaluation of Gated Recurrent Neural Networks on Sequence Modeling},
 publisher = {arXiv},
 doi = {10.48550/ARXIV.1412.3555}
}

@article{Dennis.2001,
 author = {Dennis, R. G. and Kosnik, P. E. and Gilbert, M. E. and Faulkner, J. A.},
 year = {2001},
 title = {Excitability and contractility of skeletal muscle engineered from primary cultures and cell lines},
 pages = {C288-95},
 volume = {280},
 number = {2},
 journal = {American journal of physiology. Cell physiology},
 doi = {10.1152/ajpcell.2001.280.2.C288}
}

@misc{Devlin.2018,
 author = {Devlin, Jacob and Chang, Ming-Wei and Lee, Kenton and Toutanova, Kristina},
 date = {2018},
 title = {BERT: Pre-training of Deep Bidirectional Transformers for Language Understanding},
 publisher = {arXiv},
 doi = {10.48550/ARXIV.1810.04805}
}

@misc{Dosovitskiy.2020,
 author = {Dosovitskiy, Alexey and Beyer, Lucas and Kolesnikov, Alexander and Weissenborn, Dirk and Zhai, Xiaohua and Unterthiner, Thomas and Dehghani, Mostafa and Minderer, Matthias and Heigold, Georg and Gelly, Sylvain and Uszkoreit, Jakob and Houlsby, Neil},
 date = {2020},
 title = {An Image is Worth 16x16 Words: Transformers for Image Recognition at Scale},
 publisher = {arXiv},
 doi = {10.48550/ARXIV.2010.11929}
}

@misc{Elton.2018,
 author = {Elton, Daniel C.},
 date = {2018},
 title = {Stretched Exponential Relaxation},
 publisher = {arXiv},
 doi = {10.48550/ARXIV.1808.00881}
}

@article{Flomenbom.2005,
 author = {Flomenbom, Ophir and Velonia, Kelly and Loos, Davey and Masuo, Sadahiro and Cotlet, Mircea and Engelborghs, Yves and Hofkens, Johan and Rowan, Alan E. and Nolte, Roeland J. M. and {van der Auweraer}, Mark and de Schryver, Frans C. and Klafter, Joseph},
 year = {2005},
 title = {Stretched exponential decay and correlations in the catalytic activity of fluctuating single lipase molecules},
 pages = {2368--2372},
 volume = {102},
 number = {7},
 journal = {Proceedings of the National Academy of Sciences of the United States of America},
 doi = {10.1073/pnas.0409039102}
}

@article{Goldstein.2010,
 author = {Goldstein, Jeffery A. and McNally, Elizabeth M.},
 year = {2010},
 title = {Mechanisms of muscle weakness in muscular dystrophy},
 pages = {29--34},
 volume = {136},
 number = {1},
 journal = {The Journal of general physiology},
 doi = {10.1085/jgp.201010436}
}

@misc{Gu.2023,
 author = {Gu, Albert and Dao, Tri},
 date = {2023},
 title = {Mamba: Linear-Time Sequence Modeling with Selective State Spaces},
 publisher = {arXiv},
 doi = {10.48550/ARXIV.2312.00752}
}

@misc{GuerreroViu.2021,
 author = {Guerrero-Viu, Julia and Izquierdo, Sergio and Schr{\"o}ppel, Philipp and Brox, Thomas},
 date = {2021},
 title = {Semi-Supervised Disparity Estimation with Deep Feature Reconstruction},
 publisher = {arXiv},
 doi = {10.48550/ARXIV.2106.00318}
}

@article{He.2015,
 author = {He, Kaiming and Zhang, Xiangyu and Ren, Shaoqing and Sun, Jian},
 year = {2015},
 title = {Spatial Pyramid Pooling in Deep Convolutional Networks for Visual Recognition},
 pages = {1904--1916},
 volume = {37},
 number = {9},
 journal = {IEEE transactions on pattern analysis and machine intelligence},
 doi = {10.1109/TPAMI.2015.2389824}
}

@article{Heidlauf.2016,
 author = {Heidlauf, Thomas and Klotz, Thomas and Rode, Christian and Altan, Ekin and Bleiler, Christian and Siebert, Tobias and R{\"o}hrle, Oliver},
 year = {2016},
 title = {A multi-scale continuum model of skeletal muscle mechanics predicting force enhancement based on actin-titin interaction},
 pages = {1423--1437},
 volume = {15},
 number = {6},
 journal = {Biomechanics and modeling in mechanobiology},
 doi = {10.1007/s10237-016-0772-7}
}

@article{HernandezGascon.2013,
 author = {Hern{\'a}ndez-Gasc{\'o}n, B. and Grasa, J. and Calvo, B. and Rodr{\'i}guez, J. F.},
 year = {2013},
 title = {A 3D electro-mechanical continuum model for simulating skeletal muscle contraction},
 pages = {108--118},
 volume = {335},
 journal = {Journal of theoretical biology},
 doi = {10.1016/j.jtbi.2013.06.029}
}

@article{Hill.2021,
 author = {Hill, Cameron and Brunello, Elisabetta and Fusi, Luca and Ovejero, Jes{\'u}s G. and Irving, Malcolm},
 year = {2021},
 title = {Myosin-based regulation of twitch and tetanic contractions in mammalian skeletal muscle},
 volume = {10},
 journal = {eLife},
 doi = {10.7554/eLife.68211}
}

@article{Hofemeier.2021,
 author = {Hofemeier, Arne D. and Limon, Tamara and Muenker, Till Moritz and Wallmeyer, Bernhard and Jurado, Alejandro and Afshar, Mohammad Ebrahim and Ebrahimi, Majid and Tsukanov, Roman and Oleksiievets, Nazar and Enderlein, J{\"o}rg and Gilbert, Penney M. and Betz, Timo},
 year = {2021},
 title = {Global and local tension measurements in biomimetic skeletal muscle tissues reveals early mechanical homeostasis},
 volume = {10},
 journal = {eLife},
 doi = {10.7554/eLife.60145}
}

@book{Hofemeier.2022,
 author = {Hofemeier, Arne D. and Ristau, Mariam and Luber, Mattias and Muenker, Till M. and Herkenrath, Fabian and Schmelz, Bruno and Scharfenstein, Lisa-Marie and Brandt, Matthias and Malova, Polina and Shahriyari, Mina and Rinn, Malte and Haertter, Daniel and Tiburcy, Malte and Zimmermann, Wolfram-H. and Lenz, Christof and Mamchaoui, Kamel and Bigot, Anne and Nguyen, Jo and Gilbert, Penney M. and Li{\`e}vre, Cl{\'e}mence and Dupont, Jean-Baptiste and Barrett, Philip and Mack, David and Betz, Timo},
 year = {2022},
 title = {Dystrophin is a mechanical tension modulator},
 doi = {10.1101/2022.12.23.521750}
}

@article{HUXLEY.1954,
 author = {HUXLEY, A. F. and NIEDERGERKE, R.},
 year = {1954},
 title = {Structural changes in muscle during contraction; interference microscopy of living muscle fibres},
 pages = {971--973},
 volume = {173},
 number = {4412},
 issn = {0028-0836},
 journal = {Nature},
 doi = {10.1038/173971a0}
}

@misc{Islam.2020,
 author = {Islam, Md Amirul and Jia, Sen and Bruce, Neil D. B.},
 date = {2020},
 title = {How Much Position Information Do Convolutional Neural Networks Encode?},
 publisher = {arXiv},
 doi = {10.48550/ARXIV.2001.08248}
}

@article{Istratov.1999,
 author = {Istratov, Andrei A. and Vyvenko, Oleg F.},
 year = {1999},
 title = {Exponential analysis in physical phenomena},
 pages = {1233--1257},
 volume = {70},
 number = {2},
 issn = {0034-6748},
 journal = {Review of Scientific Instruments},
 doi = {10.1063/1.1149581}
}

@article{Janssen.2025,
 author = {Janssen, M. and Chan, C.-k. and Davelaar, J. and Natarajan, I. and Olivares, H. and Ripperda, B. and R{\"o}der, J. and Rynge, M. and Wielgus, M.},
 year = {2025},
 title = {Deep learning inference with the Event Horizon Telescope},
 pages = {A60},
 volume = {698},
 issn = {0004-6361},
 journal = {Astronomy {\&} Astrophysics},
 doi = {10.1051/0004-6361/202553784}
}

@article{Jeong.2025,
 author = {Jeong, Yong Jin and Moon, Taesup},
 year = {2025},
 title = {Physics-Informed Fine-Tuning for physics discovery from random and sparse data},
 pages = {112132},
 volume = {162},
 issn = {09521976},
 journal = {Engineering Applications of Artificial Intelligence},
 doi = {10.1016/j.engappai.2025.112132}
}

@article{Johnston.2006,
 author = {Johnston, D. C.},
 year = {2006},
 title = {Stretched exponential relaxation arising from a continuous sum of exponential decays},
 volume = {74},
 number = {18},
 issn = {1098-0121},
 journal = {Physical Review B},
 doi = {10.1103/PhysRevB.74.184430}
}

@article{Juhas.2015,
 author = {Juhas, Mark and Ye, Jean and Bursac, Nenad},
 year = {2015},
 title = {Design, evaluation, and application of engineered skeletal muscle},
 pages = {81--90},
 volume = {99},
 issn = {1046-2023},
 journal = {Methods (San Diego, Calif.)},
 doi = {10.1016/j.ymeth.2015.10.002}
}

@article{Karami.2023,
 author = {Karami, Mina and Zohoor, Hassan and Calvo, Bego{\~n}a and Grasa, Jorge},
 year = {2023},
 title = {A 3D multi-scale skeletal muscle model to predict active and passive responses. Application to intra-abdominal pressure prediction},
 pages = {116222},
 volume = {415},
 issn = {00457825},
 journal = {Computer Methods in Applied Mechanics and Engineering},
 doi = {10.1016/j.cma.2023.116222}
}

@article{Kohlrausch.1854,
 author = {Kohlrausch, R.},
 year = {1854},
 title = {Theorie des elektrischen R{\"u}ckstandes in der Leidener Flasche},
 pages = {56--82},
 volume = {167},
 number = {1},
 issn = {0003-3804},
 journal = {Annalen der Physik},
 doi = {10.1002/andp.18541670103}
}

@misc{Kokhlikyan.2020,
 author = {Kokhlikyan, Narine and Miglani, Vivek and Martin, Miguel and Wang, Edward and Alsallakh, Bilal and Reynolds, Jonathan and Melnikov, Alexander and Kliushkina, Natalia and Araya, Carlos and Yan, Siqi and Reblitz-Richardson, Orion},
 date = {2020},
 title = {Captum: A unified and generic model interpretability library for PyTorch},
 publisher = {arXiv},
 doi = {10.48550/ARXIV.2009.07896}
}

@article{Korhonen.2006,
 author = {Korhonen, Marko T. and Cristea, Alexander and Al{\'e}n, Markku and H{\"a}kkinen, Keijo and Sipil{\"a}, Sarianna and Mero, Antti and Viitasalo, Jukka T. and Larsson, Lars and Suominen, Harri},
 year = {2006},
 title = {Aging, muscle fiber type, and contractile function in sprint-trained athletes},
 pages = {906--917},
 volume = {101},
 number = {3},
 issn = {8750-7587},
 journal = {Journal of applied physiology (Bethesda, Md. : 1985)},
 doi = {10.1152/japplphysiol.00299.2006}
}

@article{Larsson.2019,
 author = {Larsson, Lars and Degens, Hans and Li, Meishan and Salviati, Leonardo and Lee, Young Il and Thompson, Wesley and Kirkland, James L. and Sandri, Marco},
 year = {2019},
 title = {Sarcopenia: Aging-Related Loss of Muscle Mass and Function},
 pages = {427--511},
 volume = {99},
 number = {1},
 journal = {Physiological reviews},
 doi = {10.1152/physrev.00061.2017}
}

@article{Lee.2001,
 author = {Lee, K. C. and Siegel, J. and Webb, S. E. and L{\'e}v{\^e}que-Fort, S. and Cole, M. J. and Jones, R. and Dowling, K. and Lever, M. J. and French, P. M.},
 year = {2001},
 title = {Application of the stretched exponential function to fluorescence lifetime imaging},
 pages = {1265--1274},
 volume = {81},
 number = {3},
 issn = {0006-3495},
 journal = {Biophysical journal},
 doi = {10.1016/S0006-3495(01)75784-0}
}

@article{Lemaire.2016,
 author = {Lemaire, Koen K. and Baan, Guus C. and Jaspers, Richard T. and {van Soest}, A. J. Knoek},
 year = {2016},
 title = {Comparison of the validity of Hill and Huxley muscle-tendon complex models using experimental data obtained from rat m. soleus in situ},
 pages = {977--987},
 volume = {219},
 number = {Pt 7},
 journal = {The Journal of experimental biology},
 doi = {10.1242/jeb.128280}
}

@article{Lieber.2013,
 author = {Lieber, Richard L. and Ward, Samuel R.},
 year = {2013},
 title = {Cellular mechanisms of tissue fibrosis. 4. Structural and functional consequences of skeletal muscle fibrosis},
 pages = {C241-52},
 volume = {305},
 number = {3},
 journal = {American journal of physiology. Cell physiology},
 doi = {10.1152/ajpcell.00173.2013}
}

@article{Lindsey.1980,
 author = {Lindsey, C. P. and Patterson, G. D.},
 year = {1980},
 title = {Detailed comparison of the Williams--Watts and Cole--Davidson functions},
 pages = {3348--3357},
 volume = {73},
 number = {7},
 issn = {0021-9606},
 journal = {The Journal of Chemical Physics},
 doi = {10.1063/1.440530}
}

@article{Liu.1989,
 author = {Liu, Dong C. and Nocedal, Jorge},
 year = {1989},
 title = {On the limited memory BFGS method for large scale optimization},
 pages = {503--528},
 volume = {45},
 number = {1-3},
 issn = {0025-5610},
 journal = {Mathematical Programming},
 doi = {10.1007/BF01589116}
}

@book{Luber.2026,
 author = {Luber, Mattias and Schmelz, Bruno and Lenz, Christof and Betz, Timo},
 year = {2026},
 title = {Stretched Exponential Modeling Reveals Drug-Specific Kinetics in Human Engineered Skeletal Muscle},
 doi = {10.64898/2026.06.08.730797}
}

@article{Lukichev.2019,
 author = {Lukichev, Alexander},
 year = {2019},
 title = {Physical meaning of the stretched exponential Kohlrausch function},
 pages = {2983--2987},
 volume = {383},
 number = {24},
 issn = {03759601},
 journal = {Physics Letters A},
 doi = {10.1016/j.physleta.2019.06.029}
}

@article{Madden.2015,
 author = {Madden, Lauran and Juhas, Mark and Kraus, William E. and Truskey, George A. and Bursac, Nenad},
 year = {2015},
 title = {Bioengineered human myobundles mimic clinical responses of skeletal muscle to drugs},
 pages = {e04885},
 volume = {4},
 journal = {eLife},
 doi = {10.7554/eLife.04885}
}

@article{Mareedu.2021,
 author = {Mareedu, Satvik and Million, Emily D. and Duan, Dongsheng and Babu, Gopal J.},
 year = {2021},
 title = {Abnormal Calcium Handling in Duchenne Muscular Dystrophy: Mechanisms and Potential Therapies},
 pages = {647010},
 volume = {12},
 issn = {1664-042X},
 journal = {Frontiers in physiology},
 doi = {10.3389/fphys.2021.647010}
}

@article{Mestre.2021,
 author = {Mestre, Rafael and Garc{\'i}a, Nerea and Pati{\~n}o, Tania and Guix, Maria and Fuentes, Judith and Valerio-Santiago, Mauricio and Almi{\~n}ana, N{\'u}ria and S{\'a}nchez, Samuel},
 year = {2021},
 title = {3D-bioengineered model of human skeletal muscle tissue with phenotypic features of aging for drug testing purposes},
 volume = {13},
 number = {4},
 journal = {Biofabrication},
 doi = {10.1088/1758-5090/ac165b}
}

@article{Milicevic.2022,
 author = {Mili{\'c}evi{\'c}, Bogdan and Ivanovi{\'c}, Milo{\v{s}} and Stojanovi{\'c}, Boban and Milo{\v{s}}evi{\'c}, Miljan and Koji{\'c}, Milo{\v{s}} and Filipovi{\'c}, Nenad},
 year = {2022},
 title = {Huxley muscle model surrogates for high-speed multi-scale simulations of cardiac contraction},
 pages = {105963},
 volume = {149},
 journal = {Computers in biology and medicine},
 doi = {10.1016/j.compbiomed.2022.105963}
}

@article{MorenoJusticia.2025,
 author = {Moreno-Justicia, Roger and {van der Stede}, Thibaux and Stocks, Ben and Laitila, Jenni and Seaborne, Robert A. and {van de Loock}, Alexia and Lievens, Eline and Samodova, Diana and Mar{\'i}n-Arraiza, Leyre and Dmytriyeva, Oksana and Browaeys, Robin and {van Vossel}, Kim and Moesgaard, Lukas and Yigit, Nurten and Anckaert, Jasper and Weyns, Anneleen and {van Thienen}, Ruud and Sahl, Ronni E. and Zanoteli, Edmar and Lawlor, Michael W. and Wierer, Michael and Mestdagh, Pieter and Vandesompele, Jo and Ochala, Julien and Hostrup, Morten and Derave, Wim and Deshmukh, Atul S.},
 year = {2025},
 title = {Human skeletal muscle fiber heterogeneity beyond myosin heavy chains},
 pages = {1764},
 volume = {16},
 number = {1},
 journal = {Nature communications},
 doi = {10.1038/s41467-025-56896-6}
}

@article{Moyle.2020,
 author = {Moyle, Louise A. and Jacques, Erik and Gilbert, Penney M.},
 year = {2020},
 title = {Engineering the next generation of human skeletal muscle models: From cellular complexity to disease modeling},
 pages = {9--18},
 volume = {16},
 issn = {24684511},
 journal = {Current Opinion in Biomedical Engineering},
 doi = {10.1016/j.cobme.2020.05.006}
}

@article{Nesmith.2016,
 author = {Nesmith, Alexander P. and Wagner, Matthew A. and Pasqualini, Francesco S. and O'Connor, Blakely B. and Pincus, Mark J. and August, Paul R. and Parker, Kevin Kit},
 year = {2016},
 title = {A human in vitro model of Duchenne muscular dystrophy muscle formation and contractility},
 pages = {47--56},
 volume = {215},
 number = {1},
 journal = {The Journal of cell biology},
 doi = {10.1083/jcb.201603111}
}

@article{NicolasMetral.2001,
 author = {Nicolas-Metral, V. and Raddatz, E. and Kucera, P. and Ruegg, U. T.},
 year = {2001},
 title = {Mdx myotubes have normal excitability but show reduced contraction-relaxation dynamics},
 pages = {69--75},
 volume = {22},
 number = {1},
 issn = {0142-4319},
 journal = {Journal of muscle research and cell motility},
 doi = {10.1023/a:1010384625954}
}

@article{Oomens.2003,
 author = {Oomens, C. W. J. and Maenhout, M. and {van Oijen}, C. H. and Drost, M. R. and Baaijens, F. P.},
 year = {2003},
 title = {Finite element modelling of contracting skeletal muscle},
 pages = {1453--1460},
 volume = {358},
 number = {1437},
 issn = {0962-8436},
 journal = {Philosophical Transactions of the Royal Society of London. Series B: Biological Sciences},
 doi = {10.1098/rstb.2003.1345}
}

@article{Raissi.2019,
 author = {Raissi, M. and Perdikaris, P. and Karniadakis, G. E.},
 year = {2019},
 title = {Physics-informed neural networks: A deep learning framework for solving forward and inverse problems involving nonlinear partial differential equations},
 pages = {686--707},
 volume = {378},
 issn = {00219991},
 journal = {Journal of Computational Physics},
 doi = {10.1016/j.jcp.2018.10.045}
}

@article{Rajabian.2021,
 author = {Rajabian, Nika and Shahini, Aref and Asmani, Mohammadnabi and Vydiam, Kalyan and Choudhury, Debanik and Nguyen, Thy and Ikhapoh, Izuagie and Zhao, Ruogang and Lei, Pedro and Andreadis, Stelios T.},
 year = {2021},
 title = {Bioengineered Skeletal Muscle as a Model of Muscle Aging and Regeneration},
 pages = {74--86},
 volume = {27},
 number = {1-2},
 journal = {Tissue engineering. Part A},
 doi = {10.1089/ten.TEA.2020.0005}
}

@article{Rausch.2020,
 author = {Rausch, Martin and B{\"o}hringer, David and Steinmann, Martin and Schubert, Dirk W. and Schr{\"u}fer, Stefan and Mark, Christoph and Fabry, Ben},
 year = {2020},
 title = {Measurement of Skeletal Muscle Fiber Contractility with High-Speed Traction Microscopy},
 pages = {657--666},
 volume = {118},
 number = {3},
 issn = {0006-3495},
 journal = {Biophysical journal},
 doi = {10.1016/j.bpj.2019.12.014}
}

@article{Riddell.2023,
 author = {Riddell, Dominique O. and Hildyard, John C. W. and Harron, Rachel C. M. and Taylor-Brown, Frances and Kornegay, Joe N. and Wells, Dominic J. and Piercy, Richard J.},
 year = {2023},
 title = {Longitudinal assessment of skeletal muscle functional mechanics in the DE50-MD dog model of Duchenne muscular dystrophy},
 volume = {16},
 number = {12},
 journal = {Disease models {\&} mechanisms},
 doi = {10.1242/dmm.050395}
}

@article{Rohrle.2019,
 author = {R{\"o}hrle, Oliver and Yavuz, Utku {\c{S}}. and Klotz, Thomas and Negro, Francesco and Heidlauf, Thomas},
 year = {2019},
 title = {Multiscale modeling of the neuromuscular system: Coupling neurophysiology and skeletal muscle mechanics},
 pages = {e1457},
 volume = {11},
 number = {6},
 journal = {Wiley interdisciplinary reviews. Systems biology and medicine},
 doi = {10.1002/wsbm.1457}
}

@article{Sabett.2017,
 author = {Sabett, Christiana and Hafftka, Ariel and Sexton, Kyle and Spencer, Richard G.},
 year = {2017},
 title = {L 1 , L p , L 2 , and elastic net penalties for regularization of Gaussian component distributions in magnetic resonance relaxometry},
 volume = {46A},
 number = {2},
 issn = {1546-6086},
 journal = {Concepts in Magnetic Resonance Part A},
 doi = {10.1002/cmr.a.21427}
}

@article{Schmidt.2005,
 author = {Schmidt, Mark},
 year = {2005},
 title = {minFunc: unconstrained differentiable multivariate optimization in Matlab},
 journal = {Software available at http://www. cs. ubc. ca/ schmidtm/Software/minFunc. htm}
}

@article{Shahriyari.2022,
 author = {Shahriyari, Mina and Islam, Md Rezaul and Sakib, Sadman M. and Rinn, Malte and Rika, Anastasia and Kr{\"u}ger, Dennis and Kaurani, Lalit and Gisa, Verena and Winterhoff, Mandy and Anandakumar, Harithaa and Shomroni, Orr and Schmidt, Matthias and Salinas, Gabriela and Unger, Andreas and Linke, Wolfgang A. and Zsch{\"u}ntzsch, Jana and Schmidt, Jens and Bassel-Duby, Rhonda and Olson, Eric N. and Fischer, Andr{\'e} and Zimmermann, Wolfram-Hubertus and Tiburcy, Malte},
 year = {2022},
 title = {Engineered skeletal muscle recapitulates human muscle development, regeneration and dystrophy},
 pages = {3106--3121},
 volume = {13},
 number = {6},
 journal = {Journal of cachexia, sarcopenia and muscle},
 doi = {10.1002/jcsm.13094}
}

@article{Smith.2022,
 author = {Smith, Alec St and Luttrell, Shawn M. and Dupont, Jean-Baptiste and Gray, Kevin and Lih, Daniel and Fleming, Jacob W. and Cunningham, Nathan J. and Jepson, Sofia and Hesson, Jennifer and Mathieu, Julie and Maves, Lisa and Berry, Bonnie J. and Fisher, Elliot C. and Sniadecki, Nathan J. and Geisse, Nicholas A. and Mack, David L.},
 year = {2022},
 title = {High-throughput, real-time monitoring of engineered skeletal muscle function using magnetic sensing},
 pages = {20417314221122127},
 volume = {13},
 issn = {2041-7314},
 journal = {Journal of tissue engineering},
 doi = {10.1177/20417314221122127}
}

@article{Song.2023,
 author = {Song, Jake and Holten-Andersen, Niels and McKinley, Gareth H.},
 year = {2023},
 title = {Non-Maxwellian viscoelastic stress relaxations in soft matter},
 pages = {7885--7906},
 volume = {19},
 number = {41},
 journal = {Soft matter},
 doi = {10.1039/d3sm00736g}
}

@article{Spencer.2020,
 author = {Spencer, Richard G. and Bi, Chuan},
 year = {2020},
 title = {A Tutorial Introduction to Inverse Problems in Magnetic Resonance},
 pages = {e4315},
 volume = {33},
 number = {12},
 journal = {NMR in biomedicine},
 doi = {10.1002/nbm.4315}
}

@misc{Sundararajan.2017,
 author = {Sundararajan, Mukund and Taly, Ankur and Yan, Qiqi},
 date = {2017},
 title = {Axiomatic Attribution for Deep Networks},
 publisher = {arXiv},
 doi = {10.48550/ARXIV.1703.01365}
}

@article{Tiper.2025,
 author = {Tiper, Yekaterina and Xie, Zhuoye and Hofemeier, Arne and Lad, Heta and Luber, Mattias and Krawetz, Roman and Betz, Timo and Zimmermann, Wolfram-Hubertus and Morton, Aaron B. and Segal, Steven S. and Gilbert, Penney M.},
 year = {2025},
 title = {Optimizing electrical field stimulation parameters reveals the maximum contractile function of human skeletal muscle microtissues},
 pages = {C1160-C1176},
 volume = {328},
 number = {4},
 journal = {American journal of physiology. Cell physiology},
 doi = {10.1152/ajpcell.00308.2024}
}

@article{Tollitt.2025,
 author = {Tollitt, Benjamin R. and Jones, Samantha W. and Ohana, Jessica and Henstock, James R. and Jackson, Malcolm J. and McArdle, Anne},
 year = {2025},
 title = {A comparison of human skeletal muscle cell maturation in 2D versus 3D culture: A quantitative proteomic study},
 pages = {e70420},
 volume = {13},
 number = {12},
 journal = {Physiological reports},
 doi = {10.14814/phy2.70420}
}

@article{vanderWal.2023,
 author = {{van der Wal}, Erik and Iuliano, Alessandro and in 't Groen, Stijn L. M. and Bholasing, Anjali P. and Priesmann, Dominik and Sharma, Preeti and {den Hamer}, Bianca and Saggiomo, Vittorio and Kr{\"u}ger, Marcus and Pijnappel, W. W. M. Pim and de Greef, Jessica C.},
 year = {2023},
 title = {Highly contractile 3D tissue engineered skeletal muscles from human iPSCs reveal similarities with primary myoblast-derived tissues},
 pages = {1954--1971},
 volume = {18},
 number = {10},
 journal = {Stem cell reports},
 doi = {10.1016/j.stemcr.2023.08.014}
}

@article{vanSoest.2019,
 author = {{van Soest}, A. J. Knoek and Casius, L. J. R. and Lemaire, K. K.},
 year = {2019},
 title = {Huxley-type cross-bridge models in largeish-scale musculoskeletal models; an evaluation of computational cost},
 pages = {43--48},
 volume = {83},
 issn = {0021-9290},
 journal = {Journal of biomechanics},
 doi = {10.1016/j.jbiomech.2018.11.021}
}

@misc{Vaswani.2017,
 author = {Vaswani, Ashish and Shazeer, Noam and Parmar, Niki and Uszkoreit, Jakob and Jones, Llion and Gomez, Aidan N. and Kaiser, Lukasz and Polosukhin, Illia},
 date = {2017},
 title = {Attention Is All You Need},
 publisher = {arXiv},
 doi = {10.48550/ARXIV.1706.03762}
}

@misc{Vennerd.2021,
 author = {Venner{\o}d, Christian Bakke and Kj{\ae}rran, Adrian and Bugge, Erling Stray},
 date = {2021},
 title = {Long Short-term Memory RNN},
 publisher = {arXiv},
 doi = {10.48550/ARXIV.2105.06756}
}

@article{VesgaCastro.2022,
 author = {Vesga-Castro, Camila and Aldazabal, Javier and Vallejo-Illarramendi, Ainara and Paredes, Jacobo},
 year = {2022},
 title = {Contractile force assessment methods for in vitro skeletal muscle tissues},
 volume = {11},
 journal = {eLife},
 doi = {10.7554/eLife.77204}
}

@article{VillotaNarvaez.2022,
 author = {Villota-Narvaez, Yesid and Garz{\'o}n-Alvarado, Diego A. and R{\"o}hrle, Oliver and Ram{\'i}rez-Mart{\'i}nez, Angelica M.},
 year = {2022},
 title = {Multi-scale mechanobiological model for skeletal muscle hypertrophy},
 pages = {899784},
 volume = {13},
 issn = {1664-042X},
 journal = {Frontiers in physiology},
 doi = {10.3389/fphys.2022.899784}
}

@article{Wakeling.2023,
 author = {Wakeling, James M. and Febrer-Nafr{\'i}a, M{\'i}riam and de Groote, Friedl},
 year = {2023},
 title = {A review of the efforts to develop muscle and musculoskeletal models for biomechanics in the last 50 years},
 pages = {111657},
 volume = {155},
 issn = {0021-9290},
 journal = {Journal of biomechanics},
 doi = {10.1016/j.jbiomech.2023.111657}
}

@article{Wang.2025,
 author = {Wang, Monan and Jin, Daixin and Wang, Haibin and Xu, Xinyi and Zheng, Siyuan},
 year = {2025},
 title = {Multi-scale modeling and simulation of skeletal muscles with different fatigue degrees based on microphysiology},
 pages = {11020},
 volume = {15},
 number = {1},
 journal = {Scientific reports},
 doi = {10.1038/s41598-025-87443-4}
}

@article{Wang.2025b,
 author = {Wang, Dongmei and Wu, Jiahong and Xu, Zeyu and Jia, Jinning and Lai, Yimei and He, Zhihua},
 year = {2025},
 title = {Increased Matrix Stiffness Promotes Slow Muscle Fibre Regeneration After Skeletal Muscle Injury},
 pages = {e70423},
 volume = {29},
 number = {4},
 journal = {Journal of cellular and molecular medicine},
 doi = {10.1111/jcmm.70423}
}

@article{Wang.2025c,
 author = {Wang, Zi and Yu, Xiaotong and Wang, Chengyan and Chen, Weibo and Wang, Jiazheng and Chu, Ying-Hua and Sun, Hongwei and Li, Rushuai and Li, Peiyong and Yang, Fan and Han, Haiwei and Kang, Taishan and Lin, Jianzhong and Yang, Chen and Chang, Shufu and Shi, Zhang and Hua, Sha and Li, Yan and Hu, Juan and Zhu, Liuhong and Zhou, Jianjun and Lin, Meijing and Guo, Jiefeng and Cai, Congbo and Chen, Zhong and {Di Guo} and Yang, Guang and Qu, Xiaobo},
 year = {2025},
 title = {One for multiple: Physics-informed synthetic data boosts generalizable deep learning for fast MRI reconstruction},
 pages = {103616},
 volume = {103},
 journal = {Medical image analysis},
 doi = {10.1016/j.media.2025.103616}
}

@article{Yoshida.2025,
 author = {Yoshida, Azumi and Baba, Kazuki and Takahashi, Hironobu and Nagese, Kenichi and Shimizu, Tatsuya},
 year = {2025},
 title = {One-step fabrication of 3D-aligned human skeletal muscle tissue and measurement of contractile force for preclinical drug testing},
 pages = {101456},
 volume = {31},
 journal = {Materials today. Bio},
 doi = {10.1016/j.mtbio.2025.101456}
}

@article{Zahalak.1981,
 author = {Zahalak, George Ireneus},
 year = {1981},
 title = {A distribution-moment approximation for kinetic theories of muscular contraction},
 pages = {89--114},
 volume = {55},
 number = {1-2},
 issn = {00255564},
 journal = {Mathematical Biosciences},
 doi = {10.1016/0025-5564(81)90014-6}
}

@article{Zeng.2023,
 author = {Zeng, Wei and Hume, Donald R. and Lu, Yongtao and Fitzpatrick, Clare K. and Babcock, Colton and Myers, Casey A. and Rullkoetter, Paul J. and Shelburne, Kevin B.},
 year = {2023},
 title = {Modeling of active skeletal muscles: a 3D continuum approach incorporating multiple muscle interactions},
 pages = {1153692},
 volume = {11},
 issn = {2296-4185},
 journal = {Frontiers in bioengineering and biotechnology},
 doi = {10.3389/fbioe.2023.1153692}
}

\bigskip

\newpage

\end{document}